\documentclass{article}

\PassOptionsToPackage{square, numbers, sort&compress}{natbib}
\usepackage[preprint]{neurips_data_2024}

\usepackage[utf8]{inputenc} 
\usepackage[T1]{fontenc}    
\usepackage{hyperref}       
\usepackage{url}            
\usepackage{booktabs}       
\usepackage{amsfonts}       
\usepackage{nicefrac}       
\usepackage{microtype}      
\usepackage{xcolor}         

\usepackage{graphicx}
\usepackage{caption}
\usepackage{subcaption}
\usepackage{adjustbox}
\usepackage{amsmath}

\usepackage[most]{tcolorbox}
\tcbset{
  aibox/.style={
    width=\textwidth,
    top=0pt, bottom=0pt, left=5pt, right=5pt,
    colback=white,
    colframe=black,
    colbacktitle=black,
    enhanced,
    center,
    attach boxed title to top left={yshift=-0.1in,xshift=0.15in},
    boxed title style={boxrule=0pt,colframe=white,},
  }
}
\newtcolorbox{AIbox}[2][]{aibox,title=#2,#1}
\tcbset{
  generation/.style={
    width=\linewidth,
  }
}
\newtcolorbox{generation}[2][]{generation,title=#2,#1}

\usepackage{listings}
\newcommand{\passat}{$\text{pass}@k$~}
\newcommand{\Passat}{$\text{Pass}@k$~}

\title{HARP: A challenging human-annotated math reasoning benchmark}

\author{%
  Albert S. Yue$^1$ \\ 
  \And
  Lovish Madaan \\
  University College London \\
  \And
  Ted Moskovitz$^2$ \\
  Anthropic \\
  \AND
  DJ Strouse \\
  \And
  Aaditya K. Singh$^1$ \\
  Gatsby Unit, UCL
}

\begin{document}

\maketitle

\begin{abstract}
Math reasoning is becoming an ever increasing area of focus as we scale large language models. However, even the previously-toughest evals like MATH are now close to saturated by frontier models (90.0\% for o1-mini and 86.5\% for Gemini 1.5 Pro). We introduce HARP, Human Annotated Reasoning Problems (for Math), consisting of 5,409 problems from the US national math competitions (A(J)HSME, AMC, AIME, USA(J)MO). Of these, 4,780 have answers that are automatically check-able (with libraries such as SymPy). These problems range six difficulty levels, with frontier models performing relatively poorly on the hardest bracket of 197 problems (average accuracy 41.1\% for o1-mini, and 9.6\% for Gemini 1.5 Pro). Our dataset also features multiple choices (for 4,110 problems) and an average of two human-written, ground-truth solutions per problem, offering new avenues of research that we explore briefly. We report evaluations for many frontier models and share some interesting analyses, such as demonstrating that frontier models across families intrinsically scale their inference-time compute for more difficult problems. Finally, we open source all code used for dataset construction (including scraping) and all code for evaluation (including answer checking) to enable future research at: \url{https://github.com/aadityasingh/HARP}.
\end{abstract}

\section{Introduction}
\label{sec:intro}

\footnotetext[1]{Equal contribution}
\footnotetext[2]{Work completed while at the Gatsby Unit, UCL}

Large language models are increasingly becoming an important tool in our everyday lives \citep{chatgpt}. As part of the development process, evaluation benchmarks play a key role in measuring capabilities of these models. A capability of growing interest in recent years has been math reasoning \citep{cobbe2021gsm8k, hendrycks2021math}, as it may offer an insight to more general problem-solving abilities. However, with the increased focus on math reasoning, existing benchmarks (such as GSM8k and MATH) have become saturated, with top frontier models surpassing 90\% accuracy \citep{o1, geminiteam2024gemini15}. 

To continue evaluating models on challenging math problems, practitioners have looked to the Mathematical Association of America's AMC-series of contests, with Gemini 1.5 evaluating on AMC exams and o1 demonstrating impressive results on the AIME. However, these evaluations are often ad-hoc, without a standardized and openly accessible benchmark of prompts and answers.

In this work, we take inspiration from this move from frontier labs and introduce HARP, Human Annotated Reasoning Problems (for Math), based off the A(J)HSME, AMC, AIME, and USA(J)MO contests. Specifically, HARP consists of 5,409 problems, of which 4,780 are short answer. While the primary use of the dataset is intended as a challenging short answer math reasoning benchmark (with top models like o1-mini only achieving 41.1\% on our highest difficulty problems), our work goes beyond prior math reasoning datasets by also offering multiple choices (for 4,110 problems), and multiple human-written ground-truth solutions for problems (with an average of $\sim$2 per problem). Multiple choice evaluation offers benefits in the ease of answer checking, and offers another lens to seeing how ``general'' a model's math reasoning capabilities are (e.g., in the presence of human-written distractors \citep{writingchoices}). Multiple human-written solutions per problem offers a resource for evaluating the effect of rephrasings in context (for few-shot prompting \cite{deng2023rephrase,dubey2024llama}), and more generally, research on data diversity \citep{bukharin2024datadiversitymattersrobust} and model- vs. human-generated solutions \citep{singh2024humandatascalingselftraining,zelikman2022starbootstrappingreasoningreasoning}. In addition to these new features, we also provide human-expert-generated difficulty levels and subject labels. Beyond the dataset, we open-source all code used for scraping, processing, and evaluation (including a \texttt{sympy}-based answer checker adapted from \citet{xwin-lm}). 

Our paper is organized as follows:
\begin{itemize}
    \item Section~\ref{sec:bench} provides an overview of the dataset and its construction, including details related to scraping, processing, and annotation.
    \item Section~\ref{sec:methods} provides details on the evaluation setup used for the various analyses.
    \item Section~\ref{sec:evals} provides scores on a sampling of state-of-the-art models on HARP: o1 series \cite{o1}, Gemini series \cite{geminiteam2024gemini15}, Claude series \cite{claude3}, Llama series \cite{dubey2024llama}, and GPT-4o series \cite{gpt4o}. Given the growing importance of inference-time compute \cite{o1}, we studied the chain-of-thought-lengths of model generations, finding that most models (not just o1 series) intrinsically scale inference compute for harder problems. 
    \item Section~\ref{sec:analysis} dives deeper into some analyses on Gemini 1.5 Pro, investigating inference-time scaling using $\text{pass}@$ and $\text{maj}@$  \citep{chen2021codex, lewkowycz2022minerva, brown2024largelanguagemonkeysscaling} and considering how multiple choice evaluation differs from short answer (including examining the order of choices \cite{gupta2024changinganswerorderdecrease}).
\end{itemize}

Overall, we hope our work will provide a challenging benchmark to continue driving progress, as well as increase access to the types of problems already being adopted by many frontier labs.

\begin{figure}
    \centering
    \includegraphics[width=\linewidth]{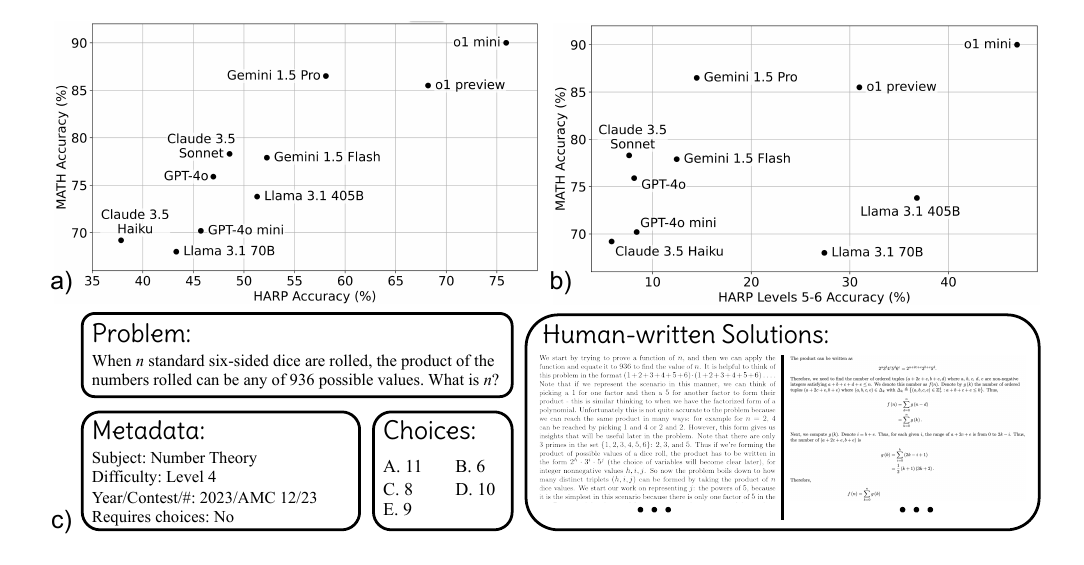}
    \vspace{-2em}
    \caption{Accuracy of various LLMs on MATH and \textbf{a)} our full dataset or \textbf{b)} our dataset restricted to the two highest difficulties. See Table~\ref{tab:overall-shortans} for numerical accuracies. We can see that improvement on MATH does not correspond to increases on the highest difficulties of HARP for most models, indicating possible overfitting to easier problems from MATH. \textbf{c)} An example problem with annotations, choices, and multiple solutions from our dataset. All 10 models we evaluated did not get this problem correct.}
    \label{fig:harp-vs-math}
    \vspace{-2.5em}
\end{figure}

\begin{figure}
    \centering
    \includegraphics[width=0.9\linewidth]{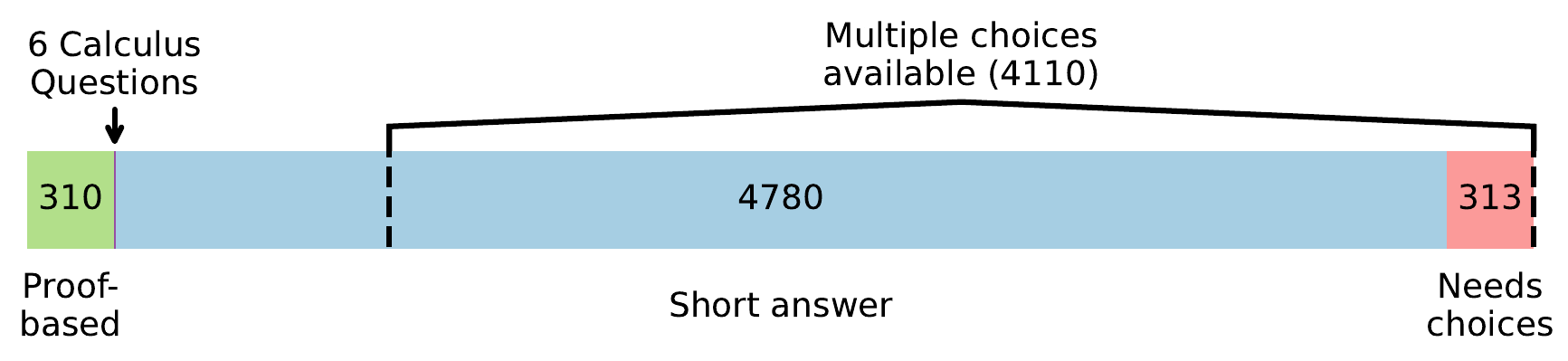}
    \includegraphics[width=0.27\linewidth]{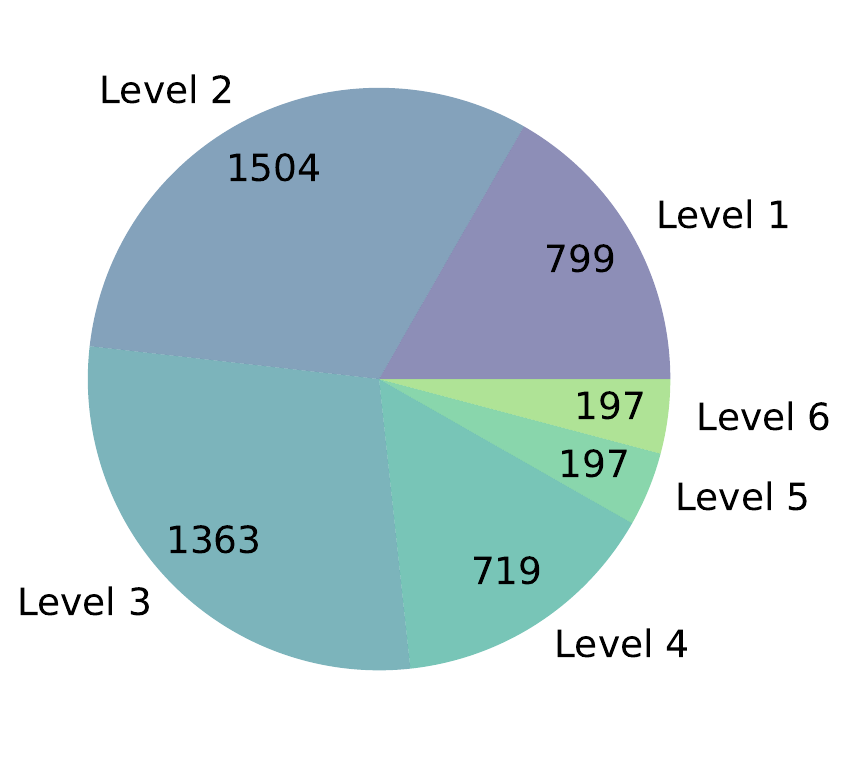}
    \includegraphics[width=0.34\linewidth]{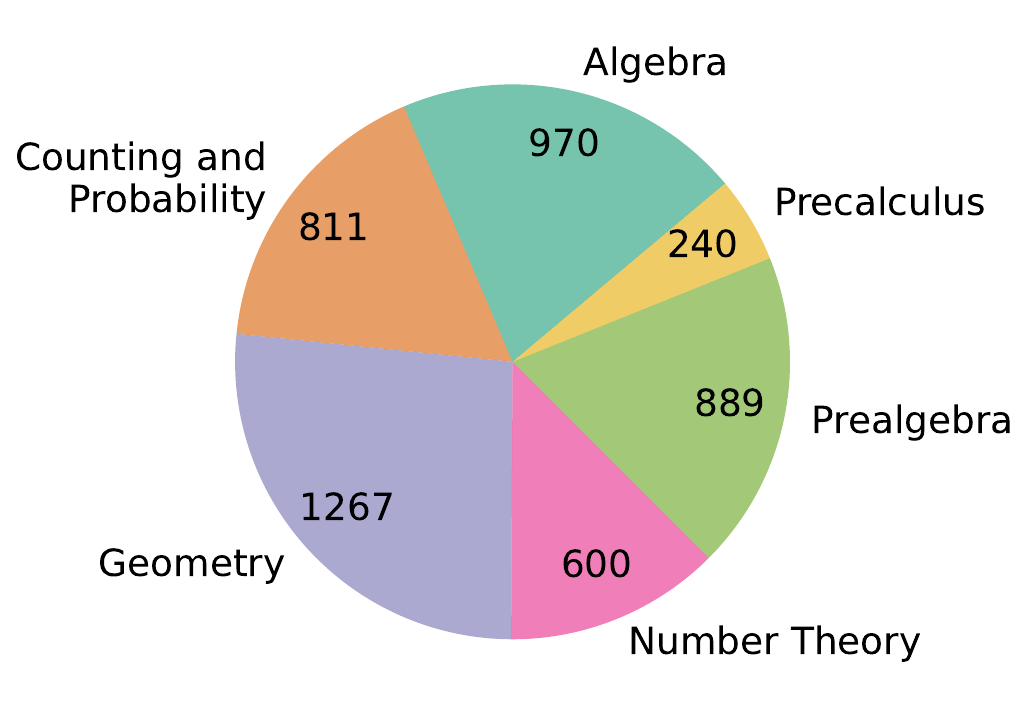}
    \includegraphics[width=0.32\linewidth]{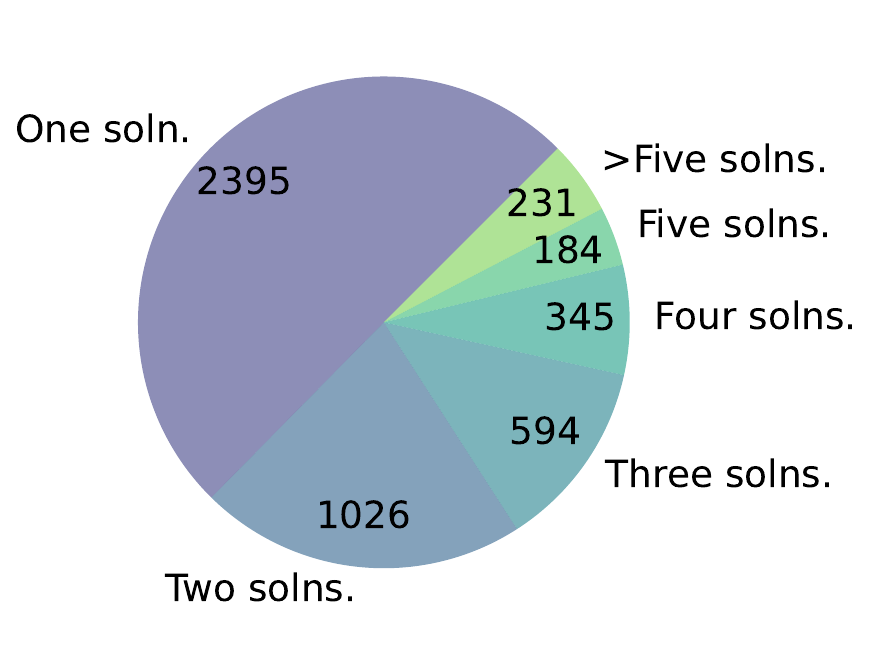}
    \caption{Dataset summary. Top row shows the breakdown of the 5,409 questions that we scraped from A(J)HSME, AMC, AIME, and USA(J)MO contests. The pie plots in the bottom row indicate the breakdown of the 4,780 short answer questions (the ``default split'' of HARP) according to difficulty level, subject, and \# of human-written ground-truth solutions.}
    \vspace{-1em}
    \label{fig:dataset}
\end{figure}

\section{Introducing HARP}
\label{sec:bench}

\subsection{Dataset summary}
\label{sec:bench:summary}

HARP consists of 5,409 problems sourced from A(J)HSME, AMC, AIME, and USA(J)MO contests that are publicly available on the AoPS Wiki.\footnote{\url{https://artofproblemsolving.com/wiki/index.php/AMC_Problems_and_Solutions}} All problems have metadata with the source contest, year, and number.
We offer a few different splits of the eval: the ``default'' split consists of the 4,780 short answer questions,\footnote{Actually, there are 4,785 short answer problems, but 5 of these problems require calculus.} with programmatically-checkable answers, and is the main data we consider for the rest of the paper (unless otherwise specified). Figure~\ref{fig:dataset} shows the composition of HARP in terms of difficulty, subject, and \# of human-written ground-truth solutions. Notably, about half the problems have more than one solution, with an average of 2.14 human-written solutions per problem and a maximum of 14 distinct human-written solutions.\footnote{Specifically, 2019 AMC8 Problem 24 has 14 solutions we were able to extract from the source: \url{https://artofproblemsolving.com/wiki/index.php/2019_AMC_8_Problems/Problem_24}} HARP also offers a multiple choice split, consisting of 4,110 problems,\footnote{Again, we exclude questions requiring calculus due to the low sample size.} which we explore in Section~\ref{sec:analysis:mc}.

\paragraph{Minimal overlap with existing benchmarks} Given that other datasets (OmniMATH \citep{gao2024omnimathuniversalolympiadlevel}, MATH \citep{hendrycks2021math}) also source from AoPS, we conducted a preliminary analysis of duplicate questions. While \citet{gao2024omnimathuniversalolympiadlevel} source from AoPS, they do not consider A(J)HSME, AMC, and AIME questions. As a result, we anticipate that our 310 Proof-based questions overlap with their dataset, but note that this is a relatively small portion of our full dataset, and completely disjoint from our ``default'' split. \citet{hendrycks2021math} also sourced from AoPS, but their dataset appears to be scraped from an older version of the Alcumus section. This results in an incomplete coverage of AMC questions, lack of choices for applicable questions, and being constrained to a single human-written solution per problem. We quantified the overlap to MATH by finding the longest prefix matches between each problem in HARP and any problem in MATH (across their train and test sets). Since our dataset construction and source differed from MATH, we chose relaxed thresholds for what should count as a duplicate: a prefix match of 60 characters or 90\% of the problem, whichever is lower. Using this procedure, we found an overlap of 790 problems, of which 781 lie in the ``default'' split of HARP (consisting of short answer problems). This  overlap is a small fraction of our dataset, and we found it useful as a means of checking consistency of our difficulty ratings and subject labels (Figure~\ref{fig:math-overlap}).\footnote{We note that any difficulty or subject labels will have some noise, which may lead to small differences between our dataset and MATH, even on duplicate problems. For example, most overlapping MATH level 5 problems are only level 3-4 in our dataset, pointing to HARP's more challenging nature. With respect to subject annotation, we see some differences in Geometry problems that we classify as Precalculus (due to the presence of trigonometric functions) and some minor discrepancy on Algebra vs. Number Theory labels. The latter is a common issue in categorization, with some human-written math contests (such as HMMT \citep{hmmt}) choosing to lump the subjects together.}

\subsection{Dataset construction}
\label{sec:bench:scraping}

We scraped a total of 6,574 problems (each corresponding to a single webpage) from the AoPS Wiki, covering all of A(J)HSME, AMC, AIME, and USA(J)MO questions from 1950 to 2024.\footnote{Our dataset does not contain AMC 10 and 12's from 2024 as these were conducted after the scrape date.} The latest scrape was run on September 20, 2024, although we did rerun scrapes for some problems after making corrections to the page for mistakes such as incorrectly transcribed answer choices.

We processed the HTML for each of these files, using image alt-text to recover LaTeX equations \cite{lewkowycz2022minerva, paster2023openwebmathopendatasethighquality, dubey2024llama}. Through a series of standardization and regular expressions, we extract the Problem and Solution blocks for each webpage. We performed some additional processing, to the best of our abilities, including:
\begin{itemize}
    \item Removal of identifiable usernames and emails.
    \item Standardization of boxed commands (to \texttt{\textbackslash boxed}, as opposed to \texttt{\textbackslash fbox}, \texttt{\textbackslash framebox}, etc.)
    \item Line-level filters to eliminate lines with hyperlinks at the ends of solutions, along with other artifacts we observed.
\end{itemize}

Following that processing, we extract problems, choices (where available), answers, and solutions. Answers are easily accessible by searching for boxed parts of solutions. For problems with multiple solutions, we enforce a consistency check across the extracted answers for each solution -- if any answer does not match, we throw out the problem. We further process solutions to remove any references to other solutions (e.g., ``as above''), as all solutions for a given problem appear on the same source webpage. While this leads to some solutions needing to be discarded, we view this as a feature of our source data, as it's unlikely that highly redundant/low diversity solutions would be present on a publicly edited wiki page. Some solutions had associated metadata in the source pages (e.g., ``Diophantine equations''), but we chose to drop these because in many cases the metadata is not meaningful (e.g., ``Faster''). Finally, we perform additional deduplication of the dataset, as problems can appear in the AMC 10 and 12 or in the USAMO and USAJMO. We detect duplicates automatically using a trie, augmented with manual deduplication for edge cases.

\subsection{Dataset annotation}
\label{sec:bench:annotation}

The constructed dataset at this stage contains 5,409 problems, with human-written choices for 4,115 problems, and an average of just over 2 human-written ground-truth solutions per problem. We assign difficulty levels to each problem, following expert guidance for the difficulty of each MAA contest.\footnote{\url{https://artofproblemsolving.com/wiki/index.php/AoPS_Wiki:Competition_ratings}}

To acquire subject labels, a human expert annotator was recruited, with extensive experience in MAA competitions.\footnote{Specifically, one of the authors of the work who had qualified for the olympiad exams and received a perfect score on the AMC10 during their high school math competition days.} They were instructed to label subjects from 8 predefined categories, with guidance provided in Appendix~\ref{appx:data:annotation}. In addition to subject annotations, we found this phase crucial to fix leftover parse issues from Section~\ref{sec:bench:scraping}. Furthermore, the annotator was instructed to annotate questions that \textit{require} choices, following the observations of \citet{myrzakhan2024openllmleaderboardmultichoiceopenstylequestions} in other natural language domains (who noted that not all multiple choice questions can be open-style). This annotation step revealed 313 problems requiring choices. To make the final ``default'' split for HARP (which consists of short answer evaluation, shown in blue in Figure~\ref{fig:dataset}), we excluded these problems, as well as the 310 proof-based problems (from the USA(J)MO) and 6 calculus questions (due to the small sample size for that subject).

\section{Evaluation methods}
\label{sec:methods}

\subsection{Models}
\label{sec:methods:models}

We evaluate ten recent models from five different model families for benchmarking HARP. Namely, Claude 3.5 \{Haiku, Sonnet\} \citep{claude35haiku,claude35sonnet}, Llama-3.1 \{70B, 405B\} \citep{dubey2024llama}, Gemini-1.5 \{Flash, Pro\} \citep{geminiteam2024gemini15}, GPT \{4o mini, 4o\} \citep{gpt4o}, o1 \{mini, preview\} \citep{o1}. We list the versions and API endpoints for each model in Appendix \ref{appx:details:models}.

We use greedy decoding ($\text{temperature} = 0$) for running most of the evaluations. We use $\text{temperature}=1$ for o1 family models as that is the only setting currently available. For the analysis in Section~\ref{sec:analysis}, we use $\text{temperature}=1$ and $\text{top\_p} = 0.95$ for all \passat evals. We set max token lengths by running models on 100 random problems from our dataset and picking values that led to less than 10 max length hits. Using this we set the max token lengths of 8192 for o1 models,\footnote{Given our results in Section~\ref{sec:evals:difficulty} -- namely that models tend to use more tokens on harder problems -- we note that o1 models may require longer context on harder problems (see Table~\ref{tab:overall-timeout}), which could lead to our numbers being underestimated. However, the significantly higher cost of these models + the lack of reproducibility (due to temperature sampling and APIs not exposing a seed at time of evaluation) made it impractical to re-run evaluations with a longer token budget for o1 models. We hope that our open-sourced data and evaluation pipeline will lead to those with more resources expanding the o1 evaluations with higher token budgets.} 4096 for Llama 3.1 405B, and 2048 for all other models.

We use and modify prompts found in their respective papers and evaluation repositories when available, and, in the case of Claude 3.5, reuse o1 prompts. The exact prompts used can be found in Appendix~\ref{appx:details:prompts}. All results presented in the main paper are zero-shot evaluations, as is most common in recent model releases, with some few-shot evaluations explored in Appendix~\ref{appx:icl}.

\subsection{Answer checker}
\label{sec:methods:scoring}

For short answer scoring, we implement a rule-based answer checker, based on the code open-sourced by Xwin-Math~\cite{xwin-lm} and expanded to handle more cases such as mixed fractions, ordered tuples, and various LaTeX commands. The checker extracts answers from model generations, performs string normalization, and then checks for equivalence as both literal strings and SymPy-parsed LaTeX math expressions. In addition, we add basic support for comparing tuples and intervals. We found that SymPy's LaTeX parser throws an error in the presence of commas, so we split the answer and evaluate each component separately when appropriate. We also found that for some complex expressions,\footnote{For example, really large numbers such as $2001^{2002^{2003}}$.} sympy number comparison hangs. We accounted for this by using a timeout of 10 seconds on all individual answer checks.

We release our code for answer checking, making it available to the community to use in future math evaluations and to further iterate upon. We also include test cases for various comparisons that the answer checker should handle. We hope that these human-annotated test cases (created in part from actual answer pairs we observed) can help serve as a resource in their own right, e.g., for comparing rule- and model-based answer checking.

For multiple choice scoring, we use slightly different prompts to encourage the model to end its response with a letter answer (still allowing for a chain-of-thought). We then extract this letter and use an exact match (ignoring case).

\begin{figure*}[t]
    \begin{subfigure}[b]{\textwidth}
        \hspace{4mm}
        \includegraphics[width=0.95\textwidth]{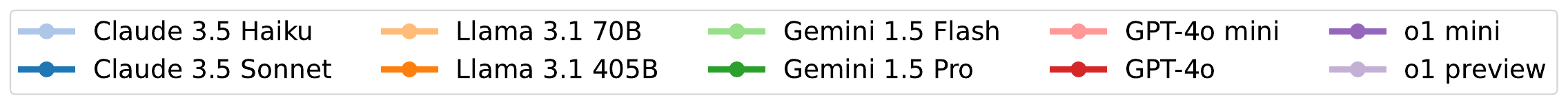}
    \end{subfigure}\\
    \begin{subfigure}[b]{\textwidth}
        \centering
        \includegraphics[height=4cm]{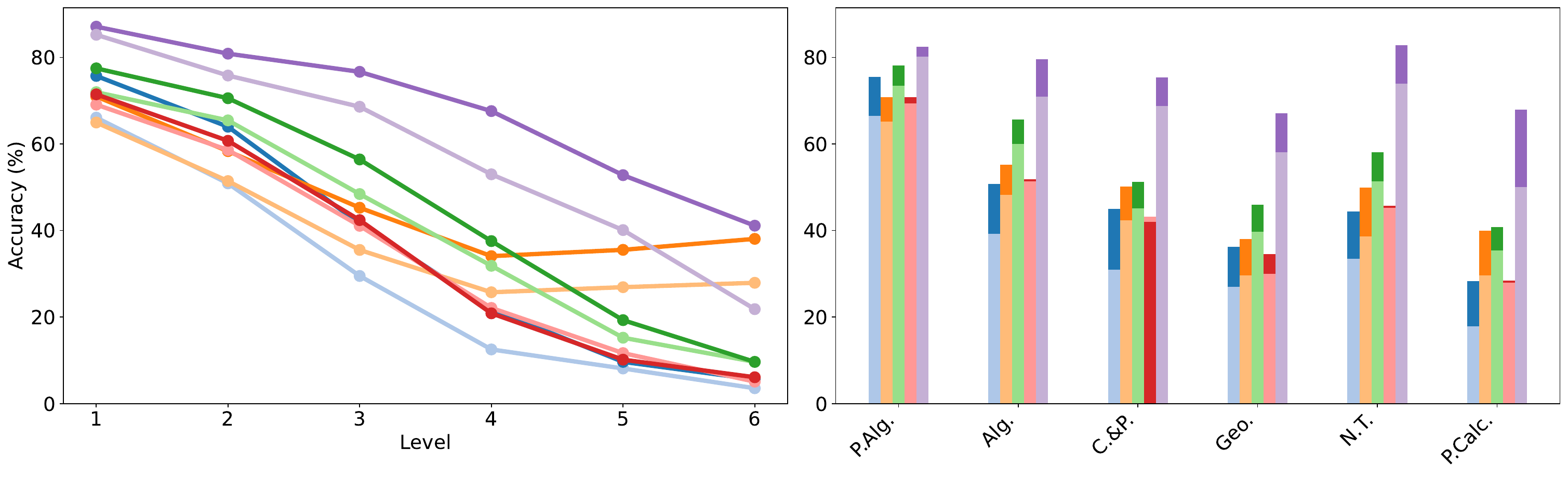}
    \end{subfigure}
    \caption{Per-difficulty (left) and Per-subject (right) accuracy of various LLMs on HARP. See Table~\ref{tab:overall-shortans} for numerical accuracies.}
    \label{fig:harp-by-group}
\end{figure*}

\section{Results across models}
\label{sec:evals}

We compare accuracies on our dataset with reported accuracy on Hendryck's MATH~\cite{hendrycks2021math} in Figure~\ref{fig:harp-vs-math}a. 
We see a positive correlation in performance between the two datasets, with lower performances across the board on HARP, leaving more room for improvement and reinforcing the notion that HARP is a challenging benchmark even for frontier LLMs. When we restrict to the two highest difficulty bins, we find that correlations weaken, with improvements in MATH accuracy corresponding to little-to-no improvement on HARP Levels 5 and 6. That said, we note that the o1 series of models show strong performance and consistent trends across MATH and HARP.

Beyond overall accuracy, we consider accuracy when problems are binned by two of the key annotations in our dataset: difficulty and subject. As seen in Figure~\ref{fig:harp-by-group}, increasing difficulty corresponds to decreased performance for nearly all models. A notable exception is the Llama series, which appear to improve slightly on the highest difficulty bins. Model generations for Llama models on this difficulty bin often ``switch'' to the right answer without any justification (see Appendix~\ref{appx:model-gens:llama-false-pos} for examples), which leads us to suspect that these problems may be contaminated in the Llama training corpus.\footnote{While we can't verify such claims without access to the corpus, we note that AoPS wiki pages may appear in public training corpora meant to emulate the Llama training corpus \cite{together2023redpajama}. We hope that the public release of our dataset prompts decontamination procedures as well as additional post-hoc contamination analyses \citep{singh2024evaluationdatacontaminationllms}.}

In terms of subjects, we find that models struggle the most with Geometry and Precalculus problems. Prealgebra problems tend to be easiest, but this specific result may be confounded by intrinsic difficulty (given that Prealgebra problems tend to be lower difficulty levels). Notably, o1 models sees improved performance across difficulties and subjects.

Detailed accuracy statistics and underlying values for all plots can be found in Appendix~\ref{appx:tables}.

\begin{figure}
    \centering
    \includegraphics[width=\linewidth]{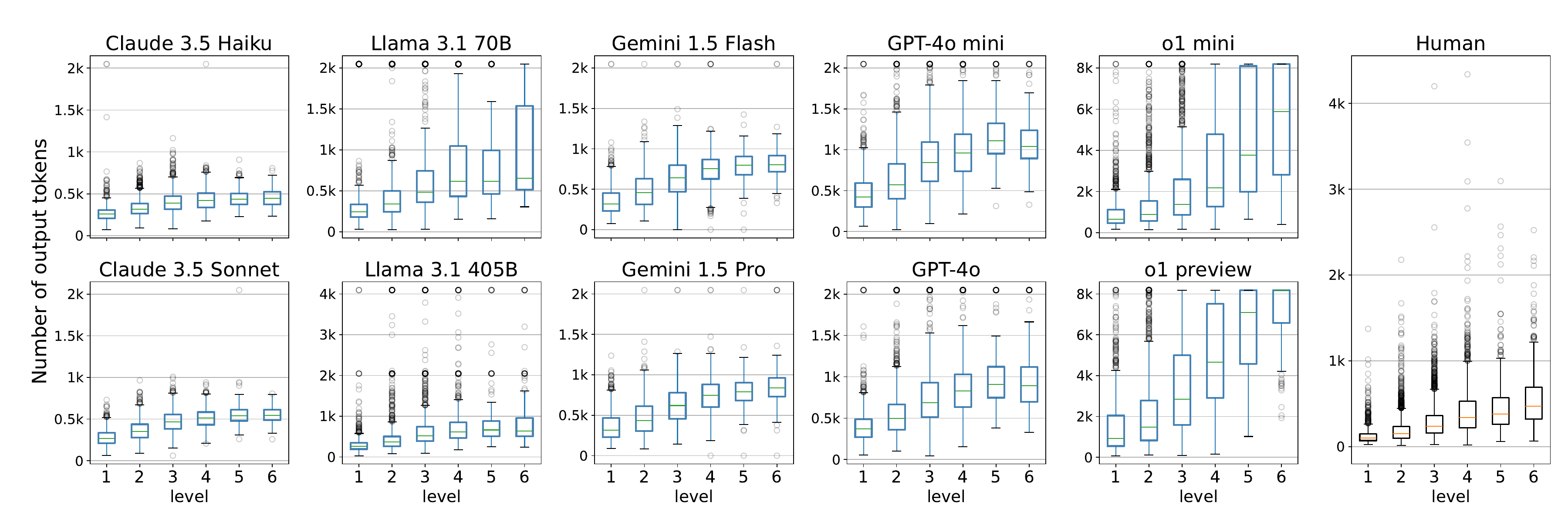}
    \caption{Distribution of number of output tokens categorized by level. The final column shows the distribution of human-written solutions. We use the GPT-4o tokenizer (via \texttt{tiktoken}) to compute the number of tokens in human-written solutions.}
    \label{fig:output-tokens}
\end{figure}

\subsection{Model chain-of-thought scales with problem difficulty}
\label{sec:evals:difficulty}

The use of more compute at inference time, in the form of chain-of-thought tokens, is becoming a direction of increasing interest \citep{o1}. Here, we test to see if models perhaps already do this implicitly: outputting chains of thought for more difficult problems. We also consider if similar effects are present in human solutions, similar to how prior work \citep{russek2022chessThinkingTime} has found human chess players to spend more time on difficult positions.

We observe models produce longer chain-of-thought reasoning when attempting\footnote{Figure \ref{fig:output-tokens-solved} shows the corresponding figure if considering problems that each model correctly solved.} harder problems (Figure~\ref{fig:output-tokens}). This behavior is consistent across all models, including o1 models where we look at the sum of reasoning and response tokens. While it may be expected for models explicitly trained to ``think'' longer \citep{o1}, this behavior is also present in models presumably not directly trained for it. That said, we do find that human ground-truths solution token length scales with difficulty, indicating that models may actually pick up on this ``how long to think'' bias of the underlying data distribution.

We also took a deeper look into cases when the models hit our set max token length. Across all models, we find that generations hit the max token length for two reasons: 1) repetition of a sequence of tokens, or 2) some form of infinitely incrementing sequence, such as the model performing casework over an infinite set of numbers. The exception are o1-series models, which use their entire token budget on unseen reasoning tokens, so we cannot verify the cause of max length hits. We show some examples in Appendix~\ref{appx:model-gens:max-length}.

\section{Deep-dive analysis}
\label{sec:analysis}

\subsection{Repeated sampling of model solutions}
\label{sec:analysis:difficulty}

\begin{figure}
    \centering
    \includegraphics[width=0.8\linewidth]{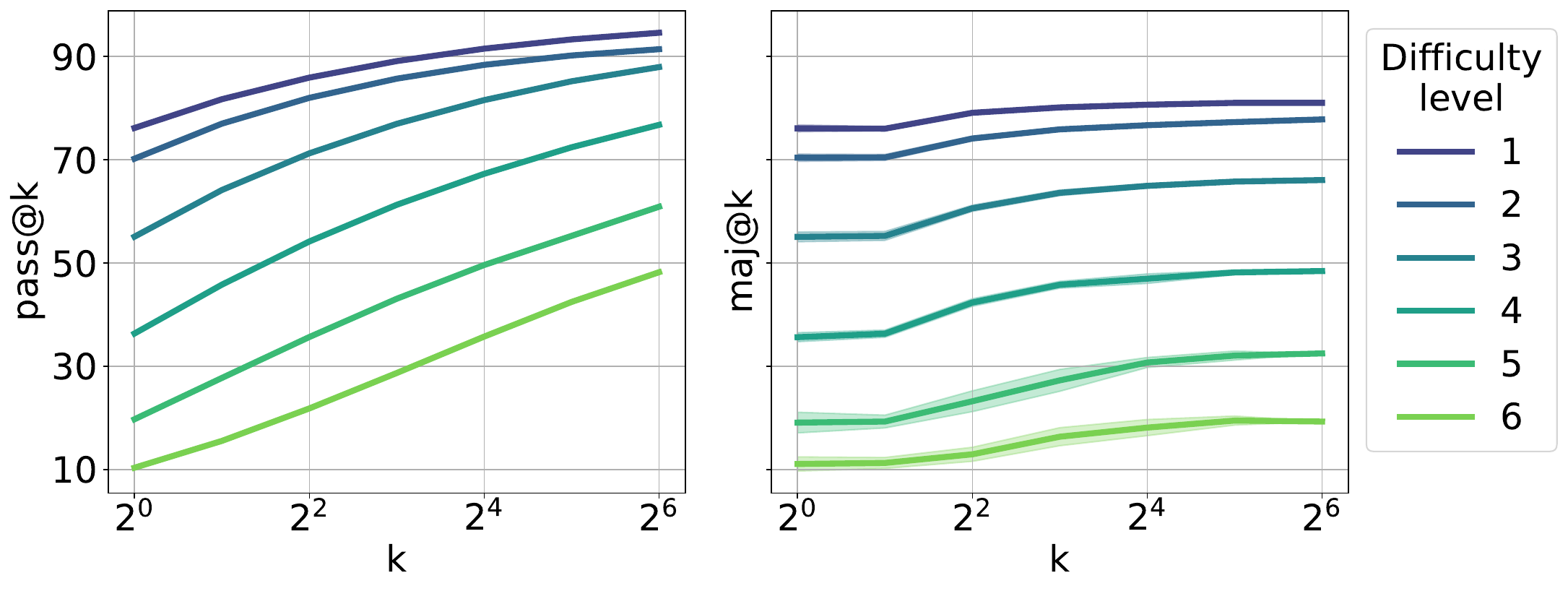}
    \caption{\Passat and $\text{maj}@k$ performance across various values of $k$ on Gemini 1.5 Pro using $\text{temperature} = 1$ and $\text{top\_p} = 0.95$. Error bars on $\text{maj}@k$ indicate 95\% confidence intervals, calculated over 5 re-orderings of samples.}
    \label{fig:pass-at-k}
\end{figure}

Recent work \cite{zelikman2022starbootstrappingreasoningreasoning,singh2024humandatascalingselftraining, hosseini2024vstartrainingverifiersselftaught, brown2024largelanguagemonkeysscaling} has focused on using inference compute to enable models to iteratively self-improve, leading to significant gains in performance. A key metric to the success of these techniques is the fraction of problems that are answered correctly at least once over $k$ samples, known as \passat. To quantify this, we sample $N$ model generations per problem, and approximate \passat for problem $i$ using the unbiased estimator $1 - \frac{\binom{N-C_i}{k}}{\binom{N}{k}}$ where $C_i$ is the number of correct samples for problem $i$ \cite{chen2021codex,yona2024inferencescalingbaseline}.
As prior work \cite{yona2024inferencescalingbaseline} has shown limits to scaling inference compute over several orders of magnitude, we set $N=64$. Overall, \passat increases from $57.3\%$ at $k=1$ to $85.7\%$ at $k=64$ (see Table~\ref{tab:passatk}). \Passat accuracy also negatively correlates with difficulty, with harder problems having lower accuracy at all values of $k$ considered (Figure~\ref{fig:pass-at-k}). 

In addition, we also compute $\text{maj}@k$ accuracy, where we take $k$ samples and mark a problem correct if the most common answer is correct.\footnote{If N answers tie as the most common, we award a score of $\frac{1}{N}$ for that problem.} Accuracy increases more slowly, only reaching $66.3\%$ at $k=64$, and we observe a similar negative correlation between $\text{maj}@k$ and difficulty (Figure~\ref{fig:pass-at-k}). Notably, benefits of consensus seem to increase with difficutly, up until the last level, indicating that models do still benefit from such inference-time techniques on medium-hard problems.

\subsection{Multiple choice evaluation}
\label{sec:analysis:mc}

We ran evaluations on the multiple choice split of HARP with Gemini 1.5 Pro. We experiment with three variations of prompting (full prompts provided in Appendix~\ref{appx:details:prompts}):
\begin{itemize}
    \item \textbf{Original text}: We use the original text in which the answer choices are presented on the AoPS Wiki. This leads to a variety of presentation styles, such as different LaTeX spacing commands ($\texttt{\textbackslash quad}$, $\texttt{\textbackslash qquad}$, etc.).
    \item \textbf{Dot format}: We present the extracted answer choices behind the letter choice and a period, i.e. $\texttt{A. <ans>}$, with each choice on its own line.
    \item \textbf{Paren. format}: Similar to dot format, we present each answer choice on its own line behind the letter choice wrapped in parentheses, i.e. $\texttt{(A) <ans>}$.
\end{itemize}
Overall, Gemini achieves an accuracy of around 80\% in all three prompting variants, performing slightly better with the paren format.
We also compare the performance of Gemini in short answer and multiple choice prompt settings. If we narrow our analysis to the intersection of the two splits, we find that short answer prompting achieves $64.2\%$ accuracy whereas multiple choice prompting achieves about $80\%$ accuracy. The improvement from short answer to multiple choice is more than the 20\% random chance of guessing the correct choice, which indicates that the availability of choices likely provides additional benefit. It's possible that answer choices helps constrain possible generations, as the model must respond with one of the five answers given as options.

Detailed accuracy statistics can be found in Appendix~\ref{appx:tables} for both the multiple choice split (Table~\ref{tab:overall-mcq}) and the overlap of the short answer and multiple choice splits (Table~\ref{tab:shortans-vs-mcq}).

\subsection{Addressing false positives through scrambled choices}
\label{sec:analysis:scramble}

\begin{figure}
    \centering
    \includegraphics[width=0.8\linewidth]{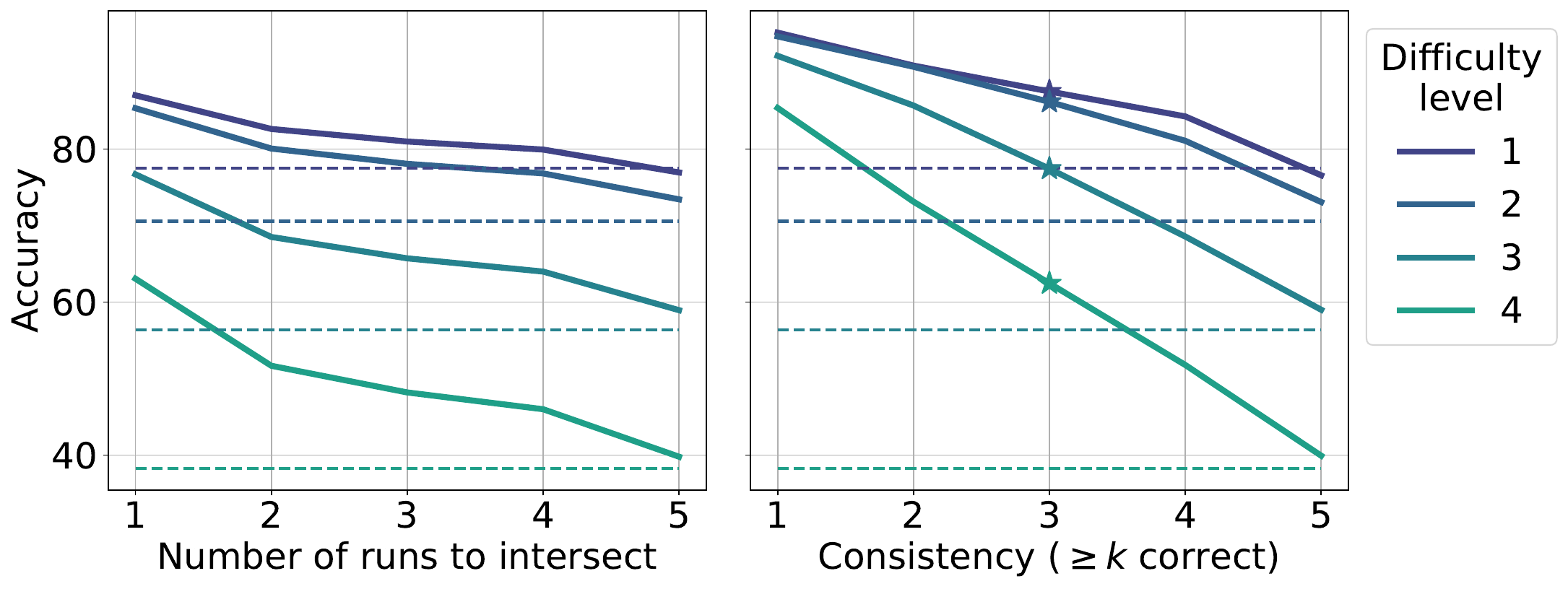}
    \caption{Accuracy across difficulties when considering mutliple choice performance over different scrambles. Dotted lines indicate evaluation accuracy when running and scoring as a short answer evaluation. (left) Accuracy when only marking problems correct if \textit{all} $x$ scrambles got the problem right (averaged over all orderings of scrambles). (right) Accuracy when only marking problems correct if \textit{at least} $x$ scrambles got the problem right (order agnostic). Stars indicate accuracies that correspond loosely to models guessing significantly above chance.}
    \label{fig:const-shuffle}
\end{figure}

For the paren. format, we create five different sets by shuffling the choices of each problem (unless it contains "All of the above" or "None of the above"). This is done to create a fairly uniform distribution for the ground truth answer across the five different choices ("A", "B", "C", "D", and "E"). We then evaluate the Gemini-1.5 Pro model to measure the consistency across the different random orderings. Figure~\ref{fig:const-shuffle} shows the consistency plot across the four difficulty levels for which we have multiple choice questions.

We find that models are not the most consistent across orderings, with performance dropping off. This inconsistency is most prevalent at higher difficulty levels. There are two ways to interpret these results. On the one hand, the lack of consistency may indicate false positives---the model guessing the correct answer choice while not actually solving the problem. To address this, one might consider restricting accuracy to problems that all runs get correct (a generalization of the pairwise metric considered in \citep{gupta2024changinganswerorderdecrease}). We present these accuracies as a function of number of orderings intersected over in Figure~\ref{fig:const-shuffle} (left). We can see that performance drops off, but is still higher than that of short answer evaluation on the same questions, even when intersecting over 5 orderings.

While the drop off may indicate false positives, it may also be induced by inherent noise in the evaluation process---for example, even a human who ``knows'' the answer to a question might not always answer correctly, depending on how the question is presented. A more statistically-minded approach may instead focus on when models perform \textit{above chance} on a problem. If we assume chance performance is 1 in 5, we can frame this as a significance test\footnote{We note that a more stringest and rigorous statistical analysis would take better account of multiple comparisons (theoretically, we're performing a test on every problem, but then we're averaging). Our goal here was to provide some intuitions, rather than a rigorous statistical analysis (which we leave to future work).} with a binomial distribution ($N=5$, $p=0.2$), in which case answering correctly in three or more orderings is significant at the $\alpha=0.05$ significance level. Per Figure~\ref{fig:const-shuffle} (left), we see that this view would imply an ever larger delta between multiple choice evaluation as compared to short answer on the corresponding problems.

Given our analysis here, we suspect the findings of Section~\ref{sec:analysis:mc} (that multiple choice evaluation leads to higher performance than short answer) are likely not solely due to the false positives.

\section{Related Work}
\label{sec:rw}

We restrict our focus in this section to other math word problem solving benchmarks, focusing on those that are also evaluated most commonly with chain-of-thought.

\paragraph{Public math benchmarks} Math reasoning has been an increasing area of focus in language model evaluation. Since the introduction of GSM8k \citep{cobbe2021gsm8k} and MATH \citep{hendrycks2021math}, abilities have progressed at a remarkable pace. \citet{paster2023hungarian} found that performance on GSM8k was saturated, with latest model improvements not corresponding to actual improvements on held-out tasks (albeit of harder difficulty). More recently, performance on MATH has also saturated, with the latest Gemini models \citep{geminiteam2024gemini15} reaching 86.5\% and OpenAI's o1-mini model hitting an impressive 90\%. In response, some new benchmarks, such as MathOdyssey \citep{fang2024mathodysseybenchmarkingmathematicalproblemsolving}, have been created but these datasets generally are limited in size (e.g., MathOdyssey is only 387 problems). This can make evaluations noisy, as results may show higher variance \citep{madaan2024quantifyingvarianceevaluationbenchmarks}. Concurrent to our work, \citet{gao2024omnimathuniversalolympiadlevel} introduce Omni-MATH, an olympiad-level math benchmark. While their approach focuses on model-generated annotations and model-based judging, we instead stick to human-annotations (for subject and difficulty) and programmatic-answer checking (via sympy). Our dataset also goes beyond these prior works by including multiple solutions and multiple choices for many problems, enabling a variety of analyses that we provide some preliminary exploration on.

\paragraph{Private math benchmarks} In response to the saturation of public benchmarks, as well as issues related to data contamination, the field has seen a proliferation of private benchmarks. These largely fall into two categories: privately sourced and publicly sourced. A great example of a privately sourced benchmark is ScaleAI's GSM1k \citep{zhang2024gsm1k}, a part of their SEAL Leaderboard \citep{scale_leaderboard}. GSM1k is a private and new version of GSM8k, created using similar guidelines. Most recently, \citet{glazer2024frontiermathbenchmarkevaluatingadvanced} introduce a challenging benchmark containing hundreds of challenging math problems, requiring many hours even for human experts to solve, and that current models struggle with (best models only achieve 2\% accuracy). On the publicly sourced side, we've seen many recent LLM releases evaluate on recent AMC and AIME problems \citep{geminiteam2024gemini15, o1}, which remain challenging for most models (e.g., Gemini 1.5 scored an 8/30 on AIME problems, though notably o1-unreleased \citet{o1} scores an impressive 11.1/15). While we acknowledge the value of such private benchmarks (especially the more challenging ones), they also increase the barrier of entry for new practitioners, or the many academic researchers working on improving math reasoning in LLMs. We hope our public dataset, that contains all of the AMC and AIME problems (which presumably make up part of these other benchmarks), will continue to drive progress in not just industrial research labs but also the academic community.

\section{Conclusion}
\label{sec:conclusion}

In this work, we introduce HARP, Human-Annotated Reasoning Problems for Math. HARP's default split consists of 4,780 short answer problems across a range of human-expert-annotated difficulties and subject categories. Recent frontier are far from saturated on our benchmark, with the best performing publicly available model at 75.9\% on the full dataset and just 41.1\% on the 197 hardest problems. Beyond the problems and answers, our dataset contains multiple choices for 4,110 problems, as well as at least 1 solution per problem (with an average of 2 solutions per problem).

\subsection{Future directions}

Our additional annotations of multiple choices and solutions also open up various interesting research questions. For example, recent work \citep{singh2024humandatascalingselftraining} found that model-generated solutions are often better than human solutions when post-training. A confound that's often hard to control is the number of solutions per problem, as it is far easier to scale for models but harder to do so for corresponding human solutions. The multiple, diverse solutions in our dataset might offer a pathway for these research directions. In terms of generating model solutions, prior work \citep{zelikman2022starbootstrappingreasoningreasoning} has attempted to use rationalizations of correct answers for post-training. In light of recent work showing the importance of model-generated negatives \citep{setlur2024rlincorrectsyntheticdata}, we hope that more datasets (specifically those intended for finetuning) include human-generated incorrect choices, which through rationalization may serve as ``harder negatives'' than random model samples with incorrect answers, as they may account for common mistakes.

Given the sourcing from AMC problems, our dataset also offers interesting work at the intersection of human and machine capabilities. For example, \citet{dasgupta2024languagemodelshumanlikecontent} find that language models often show similar biases to humans on basic reasoning tasks. Given that AMC statistics are published every year \citep{amcstats}, future work could look at whether choices that confuse human test takers more have similar effects on models.

Finally, we release all code for scraping, evaluation, and answer checking to further spur progress. We hope our main evaluation benchmark, as well as these additional public resources, continue to spur research in measuring and improving math reasoning capabilities of LLMs.

\begin{ack}
A.K.S. is funded by the Gatsby Charitable Foundation. T.M. was funded by the Gatsby Charitable Foundation during his time on this project. This work was supported by a Schmidt Science
Polymath Award to A.S., the primary supervisor of A.K.S, and the Sainsbury Wellcome Centre Core Grant from Wellcome (219627/Z/19/Z) and the
Gatsby Charitable Foundation (GAT3850).

The authors would also like to acknowledge Kira Düsterwald, Anders Andreassen, Dan Roberts, Xavier Garcia, Dieuwke Hupkes, and Andrew Lampinen for useful discussions throughout the course of this work.
\end{ack}

\bibliography{references}

\clearpage

\appendix

\section{Additional dataset details}
\label{appx:data}

\subsection{Annotation details}
\label{appx:data:annotation}

\begin{figure}
    \centering
    \fbox{\includegraphics[width=0.9\linewidth]{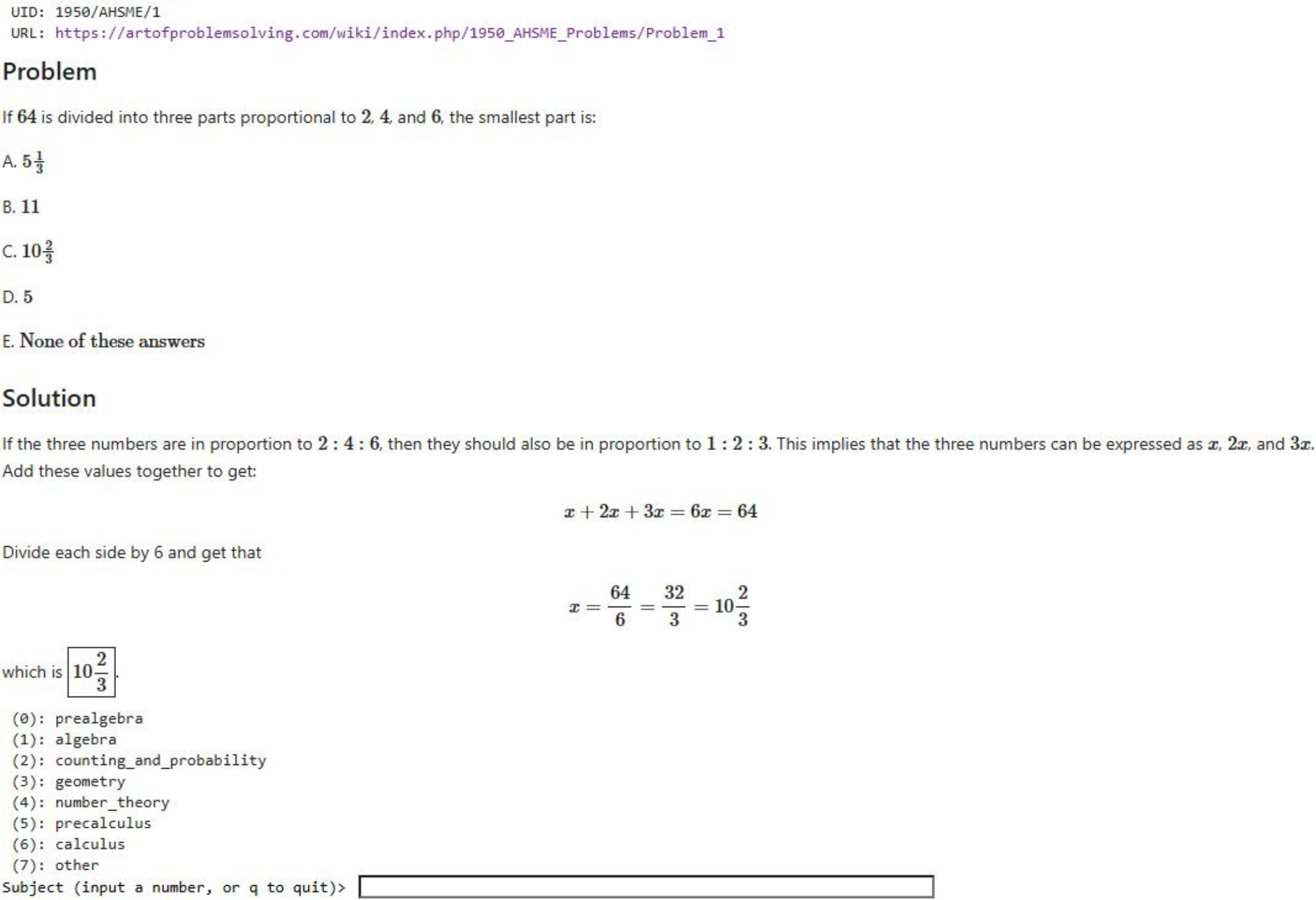}}
    \caption{An example of the panel shown for data annotation. The annotator is shown the problem number, the webpage URL, and a rendered problem and solution. We found the webpage URL to be helpful in identifying parse issues. Two input fields are give sequentially, one for subject annotation and the other for additional notes (only appearing after the subject annotation is entered).}
    \label{fig:annot-screen}
\end{figure}

We employ a human expert annotator to label subjects, problems that require choices, and other notes to make aware to the authors (e.g. noting parse issues). The annotator is provided a panel that displays the problem, first solution, and some additional metadata. They are then asked to assign a subject from 8 predetermined categories and record other notes as a comma-delimited string. An example of the panel is shown in Figure~\ref{fig:annot-screen}. When labeling the subject, we gave the annotator a table of example criteria that fit each category, which we have recreated in Table~\ref{tab:subject-instructions}.

\begin{table}
    \centering
    \begin{tabular}{ ll } 
        \hline
        Subject & Example Criteria \\
        \hline
        prealgebra & arithmetic, speed/distance, unit conversion, simple equations \\
        algebra & arithmetic/geometric sequences, systems of equations, logs, Vieta's \\
        counting\_and\_probability & stars and bars, probability, game theory \\
        geometry & areas, triangles, circles, polygons \\
        number\_theory & digit sums, modulo/remainders, prime numbers \\
        precalculus & trigonometry including law of (co)sines, complex numbers, matrices \\
        calculus & limits, continuity \\
        \hline
    \end{tabular}
    \caption{Each subject with examples given to annotator for types of problems and solutions to place into each subject category.}
    \label{tab:subject-instructions}
\end{table}

\subsection{Overlap analysis with MATH}
\label{appx:data:math-overlap}

\begin{figure}
    \centering
    \includegraphics[width=\linewidth]{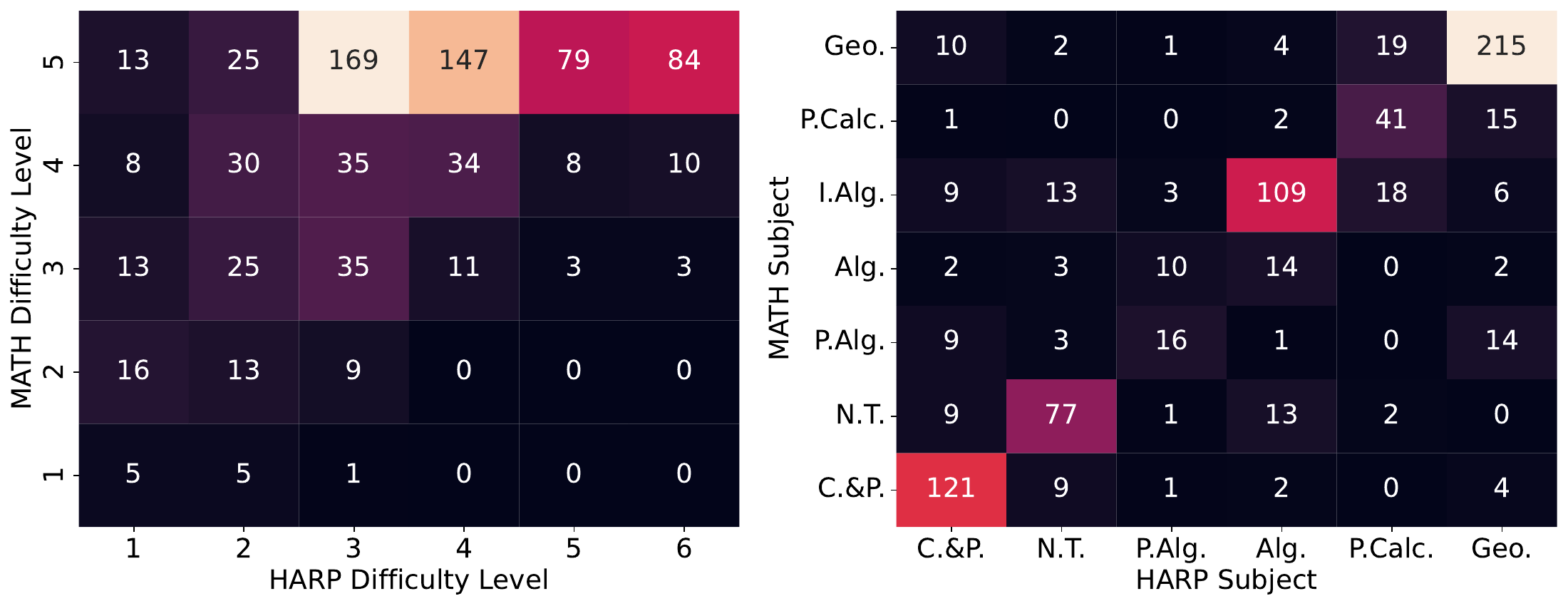}
    \caption{\textbf{a)} Difficulty level and \textbf{b)} Subject assigned to the (suspected) overlapping 781 problems in HARP and MATH. We use the ``default'' split for HARP and all problems for MATH (including train set) for calculating this figure. Notably, MATH Level 5 is often just HARP level 3. The diagonal trend for subject labels indicates high consistency between our human-expert subject annotation and that of MATH.}
    \label{fig:math-overlap}
\end{figure}

Figure~\ref{fig:math-overlap} displays a detailed breakdown of the co-occurrence of difficulty level and subject labels for problems in HARP that we suspect also appear in MATH.

\section{Additional eval details}
\label{appx:details}

\subsection{Models}
\label{appx:details:models}

We used the following APIs and model names for our experiments listed in Table \ref{tab:model-api}.
\begin{table}[h]
    \centering
    \begin{adjustbox}{width=\textwidth}
        \begin{tabular}{ lclll } 
            \hline
            Model & Max Token Length & API & API Model Name & Last Access Date \\
            \hline
            Claude 3.5 Haiku & 2048 & Anthropic & claude-3-5-haiku-20241022 & 2024-11-06 \\
            Claude 3.5 Sonnet & 2048 & Anthropic & claude-3-5-sonnet-20241022 & 2024-11-02 \\
            Llama 3.1 70B & 2048 & Together AI & meta-llama/Meta-Llama-3.1-70B-Instruct-Turbo & 2024-11-02 \\
            Llama 3.1 405B & 4096 & Together AI & meta-llama/Meta-Llama-3.1-405B-Instruct-Turbo & 2024-11-05 \\
            Gemini 1.5 Flash & 2048 & Google VertexAI & gemini-1.5-flash-002 & 2024-11-02 \\
            Gemini 1.5 Pro & 2048 & Google VertexAI & gemini-1.5-pro-002 & 2024-11-13 \\
            GPT-4o mini & 2048 & OpenAI & gpt-4o-mini-2024-07-18 & 2024-11-07 \\
            GPT-4o & 2048 & OpenAI & gpt-4o-2024-08-06 & 2024-11-07 \\
            o1 mini & 8192 & OpenAI & o1-mini-2024-09-12 & 2024-11-02 \\
            o1 preview & 8192 & OpenAI & o1-preview-2024-09-12 & 2024-11-03 \\
            \hline
        \end{tabular}
    \end{adjustbox}

    \caption{Details on models we used, such as their API specifications, max token generation length, and last access date.}
    \label{tab:model-api}
\end{table}

\subsection{Prompts}
\label{appx:details:prompts}

\subsubsection{Zero-shot evaluations across all models}

We detail the 0-shot prompt templates for the five model families below.

\begin{AIbox}{}
\textbf{Claude 3.5}\\\\
\begin{minipage}[t]{0.99\linewidth}

\textbf{System Prompt}:
\begin{lstlisting}[language=Python]
Solve the following math problem using LaTeX. The last line of your response should be of the form "Answer:\n$ANSWER" (without quotes) where $ANSWER is the answer to the problem in LaTeX. Remember to put your answer on its own line after "Answer:", and you do not need to use a \\boxed command.
\end{lstlisting}

\textbf{User prompt}:
\begin{lstlisting}[language=Python]
Problem:
{problem}
\end{lstlisting}

\textbf{Assistant prompt}:
\begin{lstlisting}[language=Python]
Solution:
\end{lstlisting}
\end{minipage}\\
\end{AIbox}

\begin{AIbox}{}
\textbf{Llama-3.1}\\\\
\begin{minipage}[t]{0.99\linewidth}

\textbf{User prompt}:
\begin{lstlisting}[language=Python]
Solve the following math problem efficiently and clearly:

- For simple problems (2 steps or fewer):
Provide a concise solution with minimal explanation.

- For complex problems (3 steps or more):
Use this step-by-step format:

## Step 1: [Concise description]
[Brief explanation and calculations]

## Step 2: [Concise description]
[Brief explanation and calculations]

...

Regardless of the approach, always conclude with:

Therefore, the final answer is: $\\boxed{{answer}}$. I hope it is correct.

Where [answer] is just the final number or expression that solves the problem.

Problem: {problem}
\end{lstlisting}
\end{minipage}\\
\end{AIbox}

\vspace{-1mm}
\begin{AIbox}{}
\textbf{Gemini 1.5}\\\\
\begin{minipage}[t]{0.99\linewidth}

\textbf{System Prompt}:
\begin{lstlisting}
You are a math expert. Solve the following math Problem, thinking step by step. At the end of the Solution, when you give your final answer, write it in the form "Final Answer: The final answer is $answer$. I hope it is correct."
\end{lstlisting}

\textbf{User prompt}:
\begin{lstlisting}[language=Python]
Problem:
{problem}
\end{lstlisting}

\textbf{Assistant prompt}:
\begin{lstlisting}[language=Python]
Solution:
\end{lstlisting}
\end{minipage}\\
\end{AIbox}

\begin{AIbox}{}
\textbf{o1}\\\\
\begin{minipage}[t]{0.99\linewidth}

\textbf{User prompt}:
\begin{lstlisting}[language=Python]
Solve the following math problem. The last line of your response should be of the form "Answer:\n$ANSWER" (without quotes) where $ANSWER is the answer to the problem. Remember to put your answer on its own line after "Answer:", and you do not need to use a \\boxed command.

{problem}
\end{lstlisting}
\end{minipage}\\
\end{AIbox}

\begin{AIbox}{}
\textbf{GPT-4o}\\\\
\begin{minipage}[t]{0.99\linewidth}

\textbf{System Prompt}:
\begin{lstlisting}[language=Python]
You are a helpful assistant.
\end{lstlisting}

\textbf{User prompt}:
\begin{lstlisting}[language=Python]
Solve the following math problem. The last line of your response should be of the form "Answer:\n$ANSWER" (without quotes) where $ANSWER is the answer to the problem. Remember to put your answer on its own line after "Answer:", and you do not need to use a \\boxed command.

{problem}
\end{lstlisting}
\end{minipage}\\
\end{AIbox}

\subsubsection{Multiple choice prompts}

\begin{AIbox}{}
\textbf{Gemini 1.5}\\\\
\begin{minipage}[t]{0.99\linewidth}

\textbf{System Prompt}:
\begin{lstlisting}
You are a math expert. Solve the following math Problem, thinking step by step. End the Solution with the final answer in the form "Final Answer: The final answer is ?. I hope it is correct.", where ? is replaced by one of the letters A, B, C, D or E.
\end{lstlisting}

\textbf{User prompt}:
\begin{lstlisting}[language=Python]
Problem:
{problem}
{choices}
\end{lstlisting}

\textbf{Assistant prompt}:
\begin{lstlisting}[language=Python]
Solution:
\end{lstlisting}
\end{minipage}\\
\end{AIbox}

\begin{AIbox}{}
\textbf{Llama-3.1}\\\\
\begin{minipage}[t]{0.99\linewidth}

\textbf{User prompt}:
\begin{lstlisting}[language=Python]
Solve the following math problem efficiently and clearly:

- For simple problems (2 steps or fewer):
Provide a concise solution with minimal explanation.

- For complex problems (3 steps or more):
Use this step-by-step format:

## Step 1: [Concise description]
[Brief explanation and calculations]

## Step 2: [Concise description]
[Brief explanation and calculations]

...

Regardless of the approach, always conclude with:

Therefore, the final answer is: ?. I hope it is correct.

Where ? is replaced by one of the letters A, B, C, D or E.

Problem: {problem}
\end{lstlisting}
\end{minipage}\\
\end{AIbox}

\subsubsection{Few-shot prompt example for Gemini}

\begin{AIbox}{}
\textbf{Gemini 1.5}\\\\
\begin{minipage}[t]{0.99\linewidth}

\textbf{System Prompt}:
\begin{lstlisting}
You are a math expert. I am going to give you a series of demonstrations of math Problems and Solutions. When you respond, respond only with the Solution of the final Problem, thinking step by step. At the end of the Solution, when you give your final answer, write it in the form "Final Answer: The final answer is $answer$. I hope it is correct."
\end{lstlisting}

$\cdots$

\textbf{User prompt}:
\begin{lstlisting}[language=Python]
Problem:
{example-problem-i}
\end{lstlisting}

\textbf{Assistant prompt}:
\begin{lstlisting}[language=Python]
Solution:
{example-solution-i}
\end{lstlisting}

$\cdots$

\textbf{User prompt}:
\begin{lstlisting}[language=Python]
Problem:
{problem}
\end{lstlisting}

\textbf{Assistant prompt}:
\begin{lstlisting}[language=Python]
Solution:
\end{lstlisting}
\end{minipage}\\
\end{AIbox}

\section{Additional results}
\label{appx:results}

\subsection{Results tables}
\label{appx:tables}
Tables corresponding to all the raw numbers for the plots in the main paper.

\begin{table}[htbp]
    \centering
    \begin{adjustbox}{width=\textwidth}
        \begin{tabular}{ |l|c|cccccc|cccccc|c| } 
            \hline
            Model & Overall Acc. & L1 & L2 & L3 & L4 & L5 & L6 & P.Alg. & Alg. & C.\&P. & Geo. & N.T. & P.Calc. & MATH Acc.\\
            \hline
            Claude 3.5 Haiku & 37.8 & 66.1 & 50.9 & 29.5 & 12.5 & 8.1 & 3.6 & 66.5 & 39.3 & 30.9 & 27.0 & 33.4 & 17.9 & 69.2 \\
            Claude 3.5 Sonnet & 48.6 & 75.7 & 64.0 & 41.6 & 21.8 & 9.6 & 5.6 & 75.5 & 50.7 & 45.0 & 36.2 & 44.4 & 28.3 & 78.3 \\
            Llama 3.1 70B & 43.3 & 65.0 & 51.4 & 35.5 & 25.7 & 26.9 & 27.9 & 65.1 & 48.2 & 42.4 & 29.7 & 38.6 & 29.6 & 68.0 \\
            Llama 3.1 405B & 51.3 & 71.0 & 58.3 & 45.3 & 34.1 & 35.5 & 38.1 & 70.9 & 55.3 & 50.1 & 38.1 & 49.9 & 40.0 & 73.8 \\
            Gemini 1.5 Flash & 52.3 & 72.0 & 65.4 & 48.4 & 31.8 & 15.2 & 9.6 & 73.5 & 60.0 & 45.1 & 39.7 & 51.4 & 35.4 & 77.9 \\
            Gemini 1.5 Pro & 58.1 & 77.5 & 70.6 & 56.4 & 37.6 & 19.3 & 9.6 & 78.2 & 65.7 & 51.2 & 45.9 & 58.1 & 40.8 & 86.5 \\
            GPT-4o mini & 45.7 & 69.1 & 58.6 & 41.1 & 22.1 & 11.7 & 5.1 & 69.4 & 51.3 & 43.2 & 30.0 & 45.3 & 27.9 & 70.2 \\
            GPT-4o & 47.0 & 71.5 & 60.7 & 42.4 & 20.9 & 10.2 & 6.1 & 70.8 & 51.3 & 42.0 & 34.5 & 45.3 & 27.9 & 75.9 \\
            o1 mini & 75.9 & 87.1 & 80.9 & 76.7 & 67.6 & 52.8 & 41.1 & 82.5 & 79.6 & 75.4 & 67.1 & 82.9 & 67.9 & 90.0* \\
            o1 preview & 68.2 & 85.2 & 75.8 & 68.6 & 53.0 & 40.1 & 21.8 & 80.2 & 70.9 & 68.7 & 58.1 & 73.9 & 50.0 & 85.5* \\
            \hline
        \end{tabular}
    \end{adjustbox}

    \caption{Accuracy results for short answer prompting on various recent LLMs. All results use zero-shot CoT prompting, over 4780 problems. MATH accuracies are as reported by their corresponding authors, and o1 models report accuracy on MATH-500.}
    \label{tab:overall-shortans}
\end{table}

\begin{table}[htbp]
    \centering
    \begin{adjustbox}{width=0.7\textwidth}
        \begin{tabular}{ |l|c|cccccc| } 
            \hline
            Model & Overall & L1 & L2 & L3 & L4 & L5 & L6 \\
            \hline
            Claude 3.5 Haiku & 0.1 & 0.3 & 0.0 & 0.0 & 0.1 & 0.0 & 0.0 \\
            Claude 3.5 Sonnet & 0.0 & 0.0 & 0.0 & 0.0 & 0.0 & 0.0 & 0.0 \\
            Llama 3.1 70B & 14.1 & 6.6 & 11.5 & 15.6 & 21.1 & 19.3 & 24.4 \\
            Llama 3.1 405B & 4.2 & 1.0 & 3.3 & 5.4 & 5.0 & 8.1 & 9.1 \\
            Gemini 1.5 Flash & 0.3 & 0.3 & 0.1 & 0.3 & 0.8 & 0.5 & 1.0 \\
            Gemini 1.5 Pro & 0.3 & 0.0 & 0.1 & 0.1 & 0.4 & 1.0 & 2.5 \\
            GPT-4o mini & 2.2 & 0.8 & 1.9 & 2.4 & 2.1 & 7.1 & 4.6 \\
            GPT-4o & 3.6 & 1.0 & 1.7 & 4.0 & 6.4 & 9.1 & 9.1 \\
            o1 mini & 6.3 & 0.5 & 1.4 & 4.5 & 11.8 & 24.9 & 39.6 \\
            o1 preview & 11.3 & 0.9 & 2.8 & 9.5 & 20.9 & 42.6 & 64.0 \\
            \hline
        \end{tabular}
    \end{adjustbox}

    \caption{Percentage of model generations that failed to complete on the short answer split. These almost always hit the set max token length, except for 4 cases in Gemini 1.5 Flash and 3 cases in Gemini 1.5 Pro where the problem was flagged as too similar to another source. These sources seemed to be sometimes correct (e.g. repost of the problem to a forum for discussion) and sometimes irrelevant (e.g. blog post about a IQ test problem). Notably, o1 series of models feature higher max-token hits on higher difficulty buckets (in line with our findings in Section~\ref{sec:evals:difficulty}). Since we count max token hits as incorrect answers (as for most models these correspond to repetition---see Figure~\ref{gen:llama-repetition},\ref{gen:llama-incrementing}), we could be underestimating performance of o1 models.}
    \label{tab:overall-timeout}
\end{table}

\begin{table}[htbp]
    \centering
    \begin{adjustbox}{width=\textwidth}
        \begin{tabular}{ |l|c|cccc|cccccc| } 
            \hline
            Prompting Method & Overall & L1 & L2 & L3 & L4 & P.Alg. & Alg. & Geo. & C.\&P. & N.T. & P.Calc. \\
            \hline
            Original text & 80.4 & 85.8 & 84.2 & 77.7 & 65.1 & 90.4 & 87.0 & 73.8 & 70.8 & 77.2 & 73.7 \\
            Dot format & 80.6 & 85.9 & 86.5 & 76.6 & 62.1 & 91.1 & 86.2 & 73.0 & 72.2 & 79.3 & 73.0 \\
            Paren format & 81.1 & 87.6 & 86.2 & 76.2 & 64.5 & 92.2 & 87.2 & 73.7 & 71.0 & 79.9 & 71.5 \\
            Paren + shuffle, run 1 & 80.7 & 87.4 & 85.4 & 76.1 & 64.7 & 92.7 & 86.4 & 72.4 & 71.3 & 78.9 & 73.0 \\
            Paren + shuffle, run 2 & 81.0 & 86.7 & 85.7 & 77.8 & 63.1 & 91.2 & 87.0 & 73.5 & 72.5 & 80.1 & 69.3 \\
            Paren + shuffle, run 3 & 80.5 & 87.2 & 85.4 & 76.2 & 62.9 & 91.4 & 85.2 & 73.3 & 72.2 & 78.9 & 71.5 \\
            Paren + shuffle, run 4 & 79.9 & 86.1 & 84.7 & 77.2 & 59.5 & 89.8 & 86.7 & 72.2 & 70.0 & 79.5 & 71.5 \\
            Paren + shuffle, run 5 & 80.3 & 87.1 & 84.6 & 76.6 & 63.7 & 91.6 & 86.9 & 72.4 & 70.0 & 78.2 & 74.5 \\
            No choices given & 20.0 & 19.9 & 19.0 & 20.2 & 22.6 & 21.8 & 20.9 & 19.0 & 17.5 & 19.5 & 21.9 \\
            \hline
        \end{tabular}
    \end{adjustbox}

    \caption{Accuracy results for multiple-choice prompting. All results are on Gemini 1.5 Pro, on 4110 problems. Shuffle runs reorder the given answer choices to a random derangement, with each run using a different derangement.}
    \label{tab:overall-mcq}
\end{table}

\begin{table}[htbp]
    \centering
    \begin{adjustbox}{width=\textwidth}
        \begin{tabular}{ |l|c|cccc|cccccc| } 
            \hline
            Prompting Method & Overall & L1 & L2 & L3 & L4 & P.Alg. & Alg. & Geo. & C.\&P. & N.T. & P.Calc. \\
            \hline
            Short answer & 64.2 & 77.5 & 70.6 & 56.3 & 38.2 & 78.1 & 70.0 & 52.4 & 57.7 & 67.3 & 48.5 \\
            Original text & 80.0 & 85.6 & 83.8 & 77.1 & 64.0 & 90.6 & 86.6 & 73.5 & 70.6 & 76.7 & 73.5 \\
            Dot format & 80.2 & 85.5 & 85.9 & 76.4 & 60.9 & 90.8 & 86.6 & 73.0 & 71.9 & 78.3 & 72.7 \\
            Paren format & 80.7 & 87.1 & 85.9 & 76.0 & 63.3 & 92.1 & 86.8 & 73.6 & 71.1 & 79.5 & 72.0 \\
            Paren + shuffle, run 1 & 80.3 & 87.0 & 85.0 & 75.5 & 64.0 & 92.7 & 86.3 & 72.2 & 71.4 & 78.1 & 72.0 \\
            Paren + shuffle, run 2 & 80.5 & 86.2 & 85.2 & 77.5 & 62.0 & 91.1 & 86.6 & 73.5 & 72.4 & 80.0 & 68.2 \\
            Paren + shuffle, run 3 & 80.1 & 86.6 & 85.0 & 75.7 & 62.4 & 91.4 & 85.1 & 73.3 & 71.9 & 78.1 & 70.5 \\
            Paren + shuffle, run 4 & 79.3 & 85.6 & 84.1 & 76.6 & 58.1 & 89.5 & 86.0 & 72.2 & 69.7 & 78.8 & 70.5 \\
            Paren + shuffle, run 5 & 80.0 & 86.6 & 84.3 & 76.2 & 62.9 & 91.4 & 86.4 & 72.5 & 70.4 & 77.9 & 73.5 \\
            No choices given & 19.5 & 19.1 & 18.5 & 20.1 & 22.3 & 21.6 & 20.3 & 18.6 & 17.4 & 18.4 & 22.0 \\
            \hline
        \end{tabular}
    \end{adjustbox}

    \caption{Accuracy results on Gemini 1.5 Pro for short answer and multiple-choice prompting, on the intersection of 3797 problems between the short answer and multiple choice splits.}
    \label{tab:shortans-vs-mcq}
\end{table}

\begin{table}[htbp]
    \centering
    \begin{adjustbox}{width=\textwidth}
        \begin{tabular}{ |l|c|cccc|cccccc| } 
            \hline
            Prompting Method & Overall & L1 & L2 & L3 & L4 & P.Alg. & Alg. & C.\&P. & Geo. & N.T. & P.Calc. \\
            \hline
            Short answer & 46.1 & 65.0 & 51.4 & 33.9 & 23.2 & 65.8 & 52.2 & 43.1 & 31.0 & 39.2 & 31.1 \\
            No choices given & 19.5 & 19.5 & 22.7 & 15.8 & 17.5 & 23.3 & 19.7 & 19.8 & 17.6 & 16.1 & 18.2 \\
            \hline
        \end{tabular}
    \end{adjustbox}

    \caption{Accuracy results on Llama 3.1 70B for short answer and multiple-choice prompting, on the intersection of 3797 problems.}
    \label{tab:llama-nochoices}
\end{table}

\begin{table}[htbp]
\centering
\begin{tabular}{lr}
\toprule
Metric & Accuracy (\%) \\
\midrule
Pass@1  & 52.41 \\
Pass@2  & 59.81 \\
Pass@4  & 65.87 \\
Pass@8  & 70.88 \\
Pass@16 & 74.96 \\
Pass@32 & 78.23 \\
Pass@64 & 80.80 \\
\bottomrule
\end{tabular}
\caption{Pass@k Accuracy Results for Gemini 1.5 Pro at temperature=1, p=0.95}
\label{tab:passatk}
\end{table}

\subsection{Additional results for output lengths}
\label{appx:results:output-lengths}
\begin{figure}
    \centering
    \includegraphics[width=\linewidth]{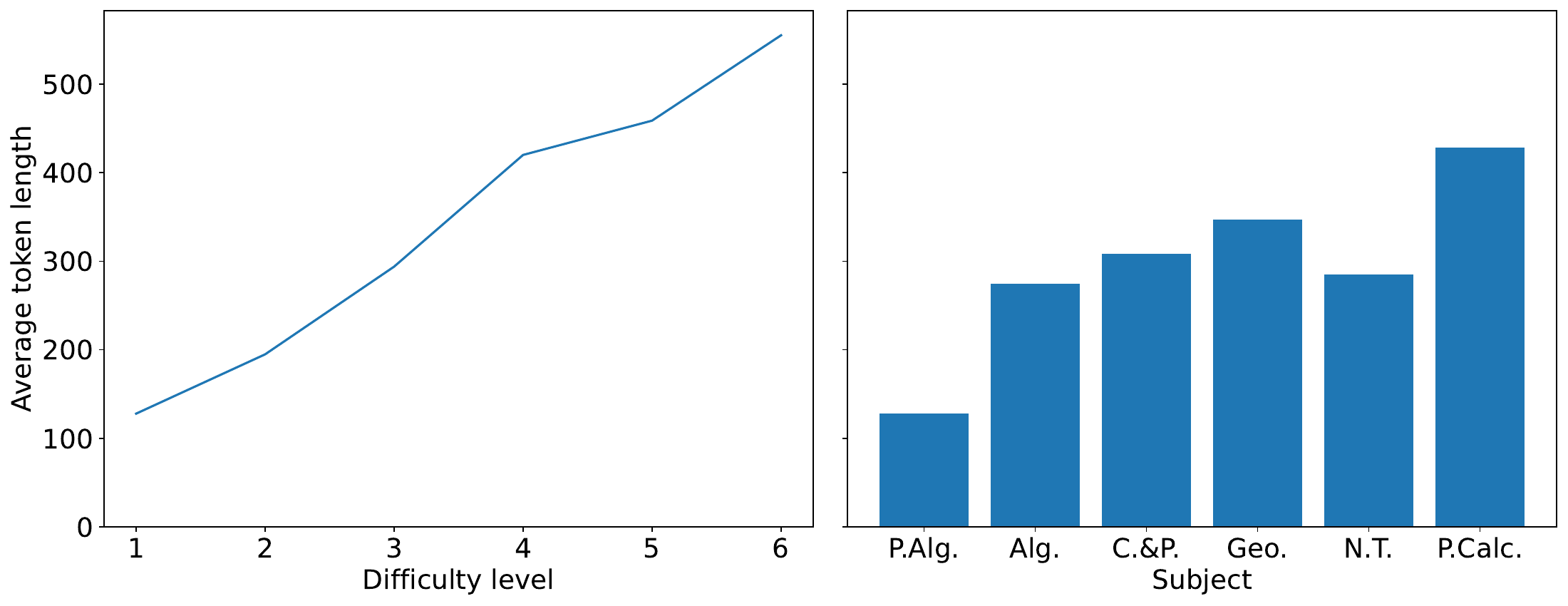}
    \caption{Average number of tokens in human-written solutions, split by level (left) or subject (right), as counted by the o1 tokenizer o200k\_base.}
    \label{fig:human-output-tokens}
\end{figure}

\begin{figure}
    \centering
    \includegraphics[width=\linewidth]{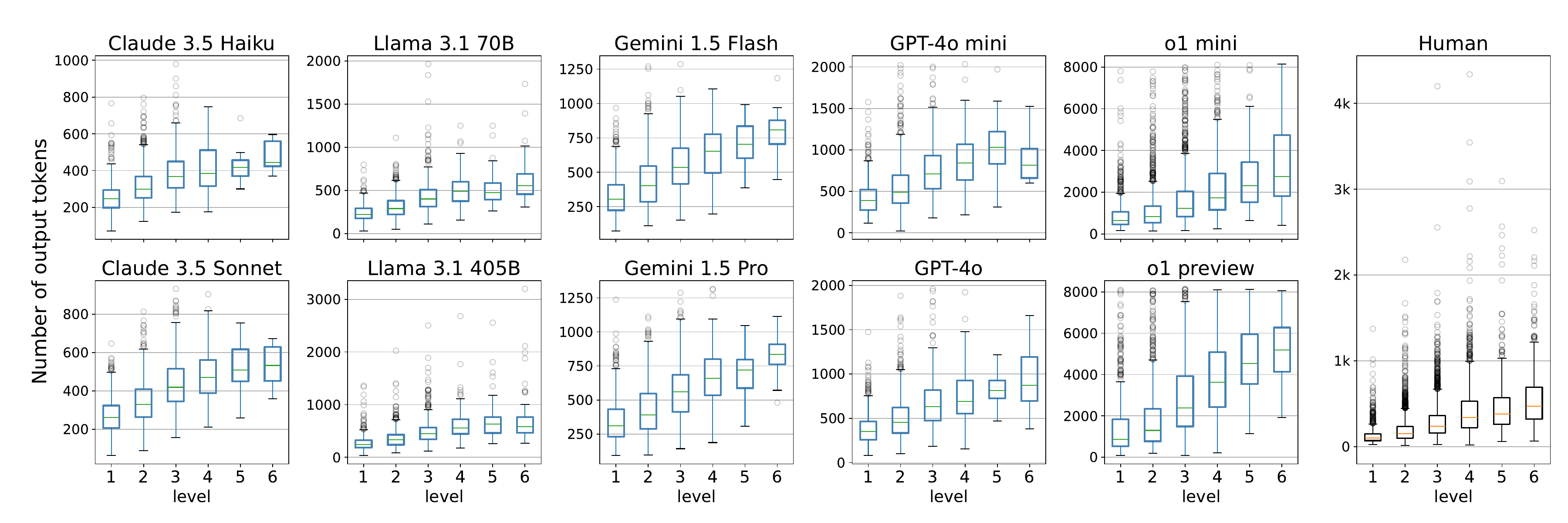}
    \caption{Distribution of number of output tokens categorized by level, restricted to problems that models solved correctly (note: this means each subfigure is made based on a different set of problems). While we find Figure~\ref{fig:output-tokens} a more compelling of the same trend, we include this plot in case it's of interest.}
    \label{fig:output-tokens-solved}
\end{figure}

We plot average number of tokens in human-written solutions in Figure~\ref{fig:human-output-tokens} and plot the distributions of output tokens on correctly solved problems in Figure~\ref{fig:output-tokens-solved}.

\subsection{Few-shot prompting for Gemini}
\label{appx:icl}

\subsubsection{Zero-shot vs 4-shot Minerva prompting}
\label{appx:icl:vs-minerva}

As the original Gemini 1.5 paper \cite{geminiteam2024gemini15} used the 4-shot Minerva prompt \cite{lewkowycz2022minerva} in benchmarking MATH, we evaluated Gemini models on the default split of HARP using both settings. We found that zero-shot prompts lead to slightly better results on both Flash and Pro, which led us to using the zero-shot prompt for future experiments. Full results are recorded in Table~\ref{tab:gemini-zero-vs-minerva}.

\begin{table}[h]
    \centering
    \begin{adjustbox}{width=\textwidth}
        \begin{tabular}{ |l|c|cccccc|cccccc| } 
            \hline
            Model & Overall & L1 & L2 & L3 & L4 & L5 & L6 & P.Alg. & Alg. & Geo. & C.\&P. & N.T. & P.Calc. \\
            \hline
            Gemini 1.5 Flash, 0-shot & 52.2 & 72.0 & 65.4 & 48.4 & 31.8 & 15.2 & 9.6 & 73.5 & 59.9 & 39.7 & 45.1 & 51.4 & 35.4 \\
            Gemini 1.5 Flash, 4-shot Minerva & 51.2 & 72.7 & 64.5 & 46.7 & 30.8 & 13.2 & 6.1 & 73.8 & 60.2 & 38.4 & 42.1 & 49.3 & 33.8 \\
            Gemini 1.5 Pro, 0-shot & 58.1 & 77.5 & 70.6 & 56.4 & 37.5 & 19.3 & 9.6 & 78.2 & 65.6 & 45.9 & 51.2 & 58.1 & 40.8 \\
            Gemini 1.5 Pro, 4-shot Minerva & 56.3 & 76.7 & 70.1 & 53.4 & 34.7 & 14.2 & 9.6 & 79.4 & 64.3 & 43.5 & 48.8 & 53.1 & 40.4 \\
            \hline
        \end{tabular}
    \end{adjustbox}

    \caption{Accuracy results on the Gemini 1.5 family, using the Gemini system and 4-shot Minerva prompts \cite{geminiteam2024gemini15,lewkowycz2022minerva} and a modified zero-shot system prompt.}
    \label{tab:gemini-zero-vs-minerva}
\end{table}

\subsubsection{Label-aligned few-shot prompts}

\begin{figure}
    \centering
    \includegraphics[width=\linewidth]{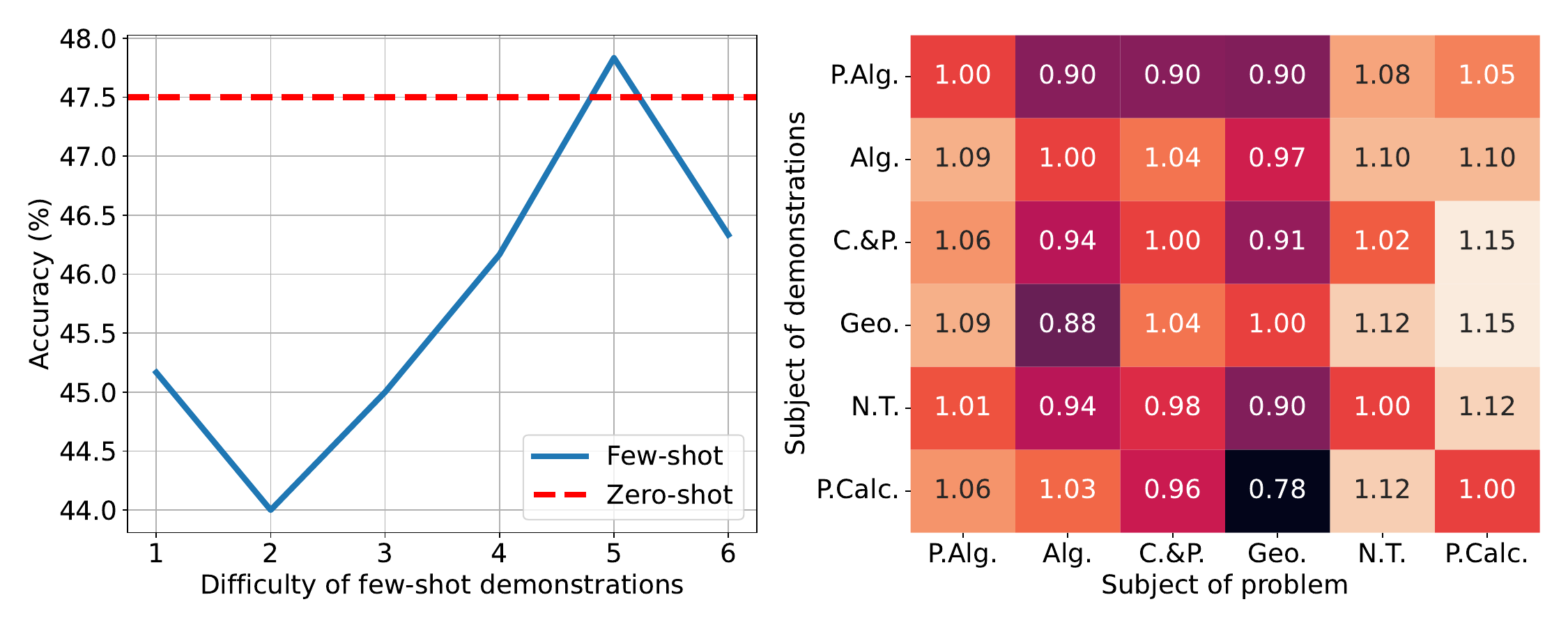}
    \caption{In-context learning experiment results. \textbf{a)} Effect of increasing difficulty in demonstrations with overall accuracy of 600 problems (100 of each of 6 difficulties), and comparison to 0-shot accuracy. \textbf{b)} Effect of demonstrations of various subjects on accuracy of 100 randomly sampled problems of a subject. Performance is normalized by column to that of demonstrations of the same subject as the problem to solve.}
    \label{fig:icl}
\end{figure}

Similar to Omni-MATH \cite{gao2024omnimathuniversalolympiadlevel}, we investigate the impact of in-context learning using problems of a given difficulty level or subject. For each, we random sample 100 problems of each difficulty or subject, and randomly select four problems of the same difficulty or subject from the remaining problems. Results are depicted in Figure~\ref{fig:icl}. We don't observe any clear trends on the impact of specific types of shots on accuracy. In particular, we don't observe a clear gain in accuracy when using harder problems as few shots, unlike in \cite{gao2024omnimathuniversalolympiadlevel}. Along with our results in Appendix~\ref{appx:icl:vs-minerva}, this may just indicate that Gemini 1.5 Pro is less sensitive to few shot examples, unlike GPT-4o, which is the model tested in \cite{gao2024omnimathuniversalolympiadlevel}.

\subsection{Select model generations}
\label{appx:model-gens}

Here are a selection of model generations that illustrate various behaviors we observed.

\subsubsection{Models repeat tokens when hitting max length}
\label{appx:model-gens:max-length}
We observe that models at temperature 0 exhibit two main behaviors that lead to reaching the max token limit, which are 1) repetition and 2) infinite incrementing, e.g. casework over an infinite set of numbers. We illustrate some examples of each in Figures~\ref{gen:gemini-repetition} and \ref{gen:gemini-incrementing} for Gemini 1.5 Pro, respectively, and 

\begin{figure}
    \centering
    \begin{generation}{}
    Let the vertices of the regular hexagon be $v_1, v_2, \dots, v_6$. Since the hexagon is centered at the origin, the vertices are given by $v_k = re^{i(2k-1)\pi/6}$ for $k = 1, 2, \dots, 6$, where $r$ is the distance from the origin to each vertex.
    The distance between opposite vertices is $2r$. The distance between opposite sides is given as 1. The distance between opposite sides is given by $r\sqrt{3}$. Thus, $r\sqrt{3} = 1$, so $r = \frac{1}{\sqrt{3}}$.
    The vertices are $v_k = \frac{1}{\sqrt{3}}e^{i(2k-1)\pi/6}$ for $k = 1, 2, \dots, 6$.
    The region $R$ is the exterior of the hexagon. The set $S$ is the image of $R$ under the transformation $w = \frac{1}{z}$.
    The transformation $w = \frac{1}{z}$ maps the exterior of the hexagon to the interior of another hexagon.
    Let $w_k = \frac{1}{v_k} = \frac{1}{\frac{1}{\sqrt{3}}e^{i(2k-1)\pi/6}} = \sqrt{3}e^{-i(2k-1)\pi/6}$ for $k = 1, 2, \dots, 6$.
    The new hexagon has vertices $w_k = \sqrt{3}e^{-i(2k-1)\pi/6}$ for $k = 1, 2, \dots, 6$.
    The distance from the origin to each vertex is $\sqrt{3}$. The side length of the new hexagon is $\sqrt{3} \cdot 2 \sin(\pi/6) = \sqrt{3}$.
    The area of the new hexagon is $6 \cdot \frac{(\sqrt{3})^2 \sqrt{3}}{4} = \frac{9\sqrt{3}}{2}$.
    The area of $S$ is the area of the hexagon with vertices $w_k$, which is $\frac{9\sqrt{3}}{2}$.
    The area of $S$ is of the form $a\pi + \sqrt{b}$. In this case, $a = 0$ and $b = \frac{81 \cdot 3}{4} = \frac{243}{4}$.
    However, $a$ and $b$ must be integers.
    The area of $S$ is $\frac{9\sqrt{3}}{2}$.
    We are given that the area of $S$ is $a\pi + \sqrt{b}$.
    Thus, $a=0$ and $\sqrt{b} = \frac{9\sqrt{3}}{2}$, so $b = \frac{81 \cdot 3}{4} = \frac{243}{4}$.
    Since $a$ and $b$ are integers, we have $a=0$ and $b = \frac{243}{4}$.
    The area of $S$ is $0\pi + \frac{9\sqrt{3}}{2}$, so $a=0$ and $b = \frac{243}{4}$.
    The area of $S$ is $\frac{9\sqrt{3}}{2} = 0\pi + \sqrt{\frac{243}{4}}$.
    Then $a=0$ and $b = \frac{243}{4}$.
    The area of $S$ is $\frac{9\sqrt{3}}{2}$, so $a=0$ and $\sqrt{b} = \frac{9\sqrt{3}}{2}$, so $b = \frac{243}{4}$.
    $a+b = 0 + \frac{243}{4}$.
    $a+b = 0 + \frac{243}{4}$.
    $a+b = 0 + \frac{243}{4}$.
    $a+b = 0 + \frac{243}{4}$.
    $a+b = 0 + \frac{243}{4}$.
    $a+b = 0 + \frac{243}{4}$.
    $a+b = 0 + \frac{243}{4}$.
    $a+b = 0 + \frac{243}{4}$.
    $a+b = 0 + \frac{243}{4}$.
    $a+b = 0 + \frac{243}{4}$.
    $a+b = 0 + \frac{243}{4}$.
    $a+b = 0 + \frac{243}{4}$.
    $a+b = 0 + \frac{243}{4}$.
    $a+b = 0 + \frac{243}{4}$.
    $a+b = 0 + \frac{243}{4}$.
    $a+b = 0 + \frac{243}{4}$.
    $a+b = 0 + \frac{243}{4}$.
    $a+b = 0 + \frac{243}{4}$.
    $a+b = 0 + \frac{243}{4}$.
    $a+b = 0 + \frac{243}{4}$.
    $a+b = 0 + \frac{243}{4}$.
    $a+b = 0 + \frac{243}{4}$.
    $a+b = 0 + \frac{243}{4}$.
    $a+b = 0 + \frac{243}{4}$.
    $a+b = 0 + \frac{243}{4}$.
    $a+b = 0 + \frac{243}{4}$.
    $a+b = 0 + \frac{243}{4}$.
    $a+b = 0 + \frac{243}{4}$.
    $a+b = 0 + \frac{243}{4}$.
    $a+b = 0 + \frac{243}{4}$.
    $a+b = 0 + \frac{243}{4}$.
    $a+b = 0 + \frac{243}{4}$.
    $a+b = 0 + \frac{243}{4}$.
    $a+b = 0 + \frac{243}{4}$.
    $a+b = 0 + \frac{243}{4}$.
    $a+b = 0 + \frac{243}{4}$.
    $a+b = 0 + \frac{243}{4}$.
    $a+b = 0 + \frac{243}{4}$.
    $a+b = 0 + \frac{243}{4}$.
    $a+b = 0 + \frac{243}{4}$.
    $a+b = 0 + \frac{243}{4}$.
    $a+b = 0 + \frac{243}{4}$.
    $a+b = 0 + \frac{243}{4}$.
    $a+b = 0 + \frac{243}{4}$.
    $a+b = 0 + \frac{243}{4}$.
    $a+b = 0 + \frac{243}{4}$.
    $a+b = 0 + \frac{243}{4}$.
    $a+b = 0 + \frac{243}{4}$.
    $a+b = 0 + \frac{243}{4}$.
    $a+b = 0 + \frac{243}{4}$.
    $a+b = 0 + \frac{243}{4}$.
    $a+b = 0 + \frac{243}{4}$.
    $a+b = 0 + \frac{243}{4}$.
    $a+b = 0 + \frac{243}{4}$.
    $a+b = 0 + \frac{243}{4}$.
    $a+b = 0 + \frac{243}{4}$.
    $a+b = 0 + \frac{243}{4}$.
    $a+b = 0 + \frac{243}{4}$.
    $a+b = 0 + \frac{243}{4}$.
    $a+b = 0 + \frac{243}{4}$.
    $a+b = 0 + \frac{243}{4}$.
    $a+b = 0 + \frac{243}{4}$.
    $a+b = 0 + \frac{243}{4}$.
    $a+b = 0 + \frac{243}{4}$.
    $a+b = 0 + \frac{243}{4}$.
    $a+b = 0 + \frac{243}{4}$.
    \$a+b = 0 +
    \end{generation}
    \caption{An example output by Gemini 1.5 Pro that hit max token length due to repetition of a sequence of tokens}
    \label{gen:gemini-repetition}
\end{figure}

\begin{figure}
    \centering
    \begin{generation}{}
        We are given that $t_1 = 1$.
        $t_2 = 1 + t_{2/2} = 1 + t_1 = 1 + 1 = 2$.
        $t_3 = \frac{1}{t_2} = \frac{1}{2}$.
        $t_4 = 1 + t_{4/2} = 1 + t_2 = 1 + 2 = 3$.
        $t_5 = \frac{1}{t_4} = \frac{1}{3}$.
        $t_6 = 1 + t_{6/2} = 1 + t_3 = 1 + \frac{1}{2} = \frac{3}{2}$.
        $t_7 = \frac{1}{t_6} = \frac{1}{3/2} = \frac{2}{3}$.
        $t_8 = 1 + t_{8/2} = 1 + t_4 = 1 + 3 = 4$.
        $t_9 = \frac{1}{t_8} = \frac{1}{4}$.
        $t_{10} = 1 + t_{10/2} = 1 + t_5 = 1 + \frac{1}{3} = \frac{4}{3}$.
        $t_{11} = \frac{1}{t_{10}} = \frac{1}{4/3} = \frac{3}{4}$.
        $t_{12} = 1 + t_{12/2} = 1 + t_6 = 1 + \frac{3}{2} = \frac{5}{2}$.
        $t_{13} = \frac{1}{t_{12}} = \frac{1}{5/2} = \frac{2}{5}$.
        $t_{14} = 1 + t_{14/2} = 1 + t_7 = 1 + \frac{2}{3} = \frac{5}{3}$.
        $t_{15} = \frac{1}{t_{14}} = \frac{1}{5/3} = \frac{3}{5}$.
        $t_{16} = 1 + t_{16/2} = 1 + t_8 = 1 + 4 = 5$.
        $t_{17} = \frac{1}{t_{16}} = \frac{1}{5}$.
        $t_{18} = 1 + t_{18/2} = 1 + t_9 = 1 + \frac{1}{4} = \frac{5}{4}$.
        $t_{19} = \frac{1}{t_{18}} = \frac{1}{5/4} = \frac{4}{5}$.
        $t_{20} = 1 + t_{20/2} = 1 + t_{10} = 1 + \frac{4}{3} = \frac{7}{3}$.
        $t_{21} = \frac{1}{t_{20}} = \frac{1}{7/3} = \frac{3}{7}$.
        $t_{22} = 1 + t_{22/2} = 1 + t_{11} = 1 + \frac{3}{4} = \frac{7}{4}$.
        $t_{23} = \frac{1}{t_{22}} = \frac{1}{7/4} = \frac{4}{7}$.
        $t_{24} = 1 + t_{24/2} = 1 + t_{12} = 1 + \frac{5}{2} = \frac{7}{2}$.
        $t_{25} = \frac{1}{t_{24}} = \frac{1}{7/2} = \frac{2}{7}$.
        $t_{26} = 1 + t_{26/2} = 1 + t_{13} = 1 + \frac{2}{5} = \frac{7}{5}$.
        $t_{27} = \frac{1}{t_{26}} = \frac{1}{7/5} = \frac{5}{7}$.
        $t_{86} = 1 + t_{43} = 1 + \frac{1}{t_{42}} = 1 + \frac{1}{1 + t_{21}} = 1 + \frac{1}{1 + \frac{3}{7}} = 1 + \frac{1}{\frac{10}{7}} = 1 + \frac{7}{10} = \frac{17}{10}$.
        $t_{87} = \frac{1}{t_{86}} = \frac{10}{17}$.
        $t_{174} = 1 + t_{87} = 1 + \frac{10}{17} = \frac{27}{17}$.
        $t_{348} = 1 + t_{174} = 1 + \frac{27}{17} = \frac{44}{17}$.
        $t_{696} = 1 + t_{348} = 1 + \frac{44}{17} = \frac{61}{17}$.
        $t_{1392} = 1 + t_{696} = 1 + \frac{61}{17} = \frac{78}{17}$.
        $t_{2784} = 1 + t_{1392} = 1 + \frac{78}{17} = \frac{95}{17}$.
        $t_{5568} = 1 + t_{2784} = 1 + \frac{95}{17} = \frac{112}{17}$.
        $t_{11136} = 1 + t_{5568} = 1 + \frac{112}{17} = \frac{129}{17}$.
        $t_{22272} = 1 + t_{11136} = 1 + \frac{129}{17} = \frac{146}{17}$.
        $t_{44544} = 1 + t_{22272} = 1 + \frac{146}{17} = \frac{163}{17}$.
        $t_{89088} = 1 + t_{44544} = 1 + \frac{163}{17} = \frac{180}{17}$.
        $t_{178176} = 1 + t_{89088} = 1 + \frac{180}{17} = \frac{197}{17}$.
        $t_{356352} = 1 + t_{178176} = 1 + \frac{197}{17} = \frac{214}{17}$.
        $t_{712704} = 1 + t_{356352} = 1 + \frac{214}{17} = \frac{231}{17}$.
        $t_{1425408} = 1 + t_{712704} = 1 + \frac{231}{17} = \frac{248}{17}$.
        $t_{2850816} = 1 + t_{1425408} = 1 + \frac{248}{17} = \frac{265}{17}$.
        $t_{5701632} = 1 + t_{2850816} = 1 + \frac{265}{17} = \frac{282}{17}$.
        $t_{11403264} = 1 + t_{5701632} = 1 + \frac{282}{17} = \frac{299}{17}$.
        $t_{22806528} = 1 + t_{11403264} = 1 + \frac{299}{17} = \frac{316}{17}$.
        $t_{45613056} = 1 + t_{22806528} = 1 + \frac{316}{17} = \frac{333}{17}$.
        $t_{91226112} = 1 + t_{45613056} = 1 + \frac{333}{17} = \frac{350}{17}$.
        \$t\_\{1824522
    \end{generation}
    \caption{An example output by Gemini 1.5 Pro that hit max token length due to incrementing infinitely.}
    \label{gen:gemini-incrementing}
\end{figure}

\begin{figure}
    \centering
    \begin{generation}{}
        \#\# Step 1: Add the given equations to eliminate the variable xyz
        Adding the given equations, we get $x^3 + y^3 + z^3 - 3xyz = 28$. We can rewrite this as $x^3 + y^3 + z^3 = 28 + 3xyz$.
        
        \#\# Step 2: Use the factorization of the sum of cubes to simplify the equation
        Using the factorization $x^3 + y^3 + z^3 - 3xyz = (x + y + z)(x^2 + y^2 + z^2 - xy - yz - zx)$, we can rewrite the equation as $(x + y + z)(x^2 + y^2 + z^2 - xy - yz - zx) = 28$.
        
        \#\# Step 3: Apply the AM-GM inequality to find the maximum value of the sum of cubes
        Applying the AM-GM inequality to the equation, we get $(x + y + z)(x^2 + y^2 + z^2 - xy - yz - zx) \leq \frac{(x + y + z)^3}{3}$. Since $(x + y + z)(x^2 + y^2 + z^2 - xy - yz - zx) = 28$, we have $\frac{(x + y + z)^3}{3} \geq 28$.
        
        \#\# Step 4: Find the maximum value of the sum of cubes
        From the inequality in Step 3, we can find the maximum value of $(x + y + z)^3$. Since $\frac{(x + y + z)^3}{3} \geq 28$, we have $(x + y + z)^3 \geq 84$. Therefore, the maximum value of $x^3 + y^3 + z^3$ is $28 + 3xyz = 28 + 3 \cdot \frac{(x + y + z)^3}{3} = 28 + (x + y + z)^3 \leq 28 + 84 = 112$.
        
        \#\# Step 5: Find the greatest possible value of $a^3 + b^3 + c^3$
        Since the maximum value of $x^3 + y^3 + z^3$ is 112, the greatest possible value of $a^3 + b^3 + c^3$ is $\frac{112}{3} \cdot \frac{3}{2} = \frac{336}{6} = \frac{168}{3} = 56$ is incorrect. However, $\frac{112}{3} \cdot \frac{3}{2} = \frac{336}{6} = \frac{168}{3} = 56$ is incorrect. However, $\frac{112}{3} \cdot \frac{3}{2} = \frac{336}{6} = \frac{168}{3} = 56$ is incorrect. However, $\frac{112}{3} \cdot \frac{3}{2} = \frac{336}{6} = \frac{168}{3} = 56$ is incorrect. However, $\frac{112}{3} \cdot \frac{3}{2} = \frac{336}{6} = \frac{168}{3} = 56$ is incorrect. However, $\frac{112}{3} \cdot \frac{3}{2} = \frac{336}{6} = \frac{168}{3} = 56$ is incorrect. However, $\frac{112}{3} \cdot \frac{3}{2} = \frac{336}{6} = \frac{168}{3} = 56$ is incorrect. However, $\frac{112}{3} \cdot \frac{3}{2} = \frac{336}{6} = \frac{168}{3} = 56$ is incorrect. However, $\frac{112}{3} \cdot \frac{3}{2} = \frac{336}{6} = \frac{168}{3} = 56$ is incorrect. However, $\frac{112}{3} \cdot \frac{3}{2} = \frac{336}{6} = \frac{168}{3} = 56$ is incorrect. However, $\frac{112}{3} \cdot \frac{3}{2} = \frac{336}{6} = \frac{168}{3} = 56$ is incorrect. However, $\frac{112}{3} \cdot \frac{3}{2} = \frac{336}{6} = \frac{168}{3} = 56$ is incorrect. However, $\frac{112}{3} \cdot \frac{3}{2} = \frac{336}{6} = \frac{168}{3} = 56$ is incorrect. However, $\frac{112}{3} \cdot \frac{3}{2} = \frac{336}{6} = \frac{168}{3} = 56$ is incorrect. However, $\frac{112}{3} \cdot \frac{3}{2} = \frac{336}{6} = \frac{168}{3} = 56$ is incorrect. However, $\frac{112}{3} \cdot \frac{3}{2} = \frac{336}{6} = \frac{168}{3} = 56$ is incorrect. However, $\frac{112}{3} \cdot \frac{3}{2} = \frac{336}{6} = \frac{168}{3} = 56$ is incorrect. However, $\frac{112}{3} \cdot \frac{3}{2} = \frac{336}{6} = \frac{168}{3} = 56$ is incorrect. However, $\frac{112}{3} \cdot \frac{3}{2} = \frac{336}{6} = \frac{168}{3} = 56$ is incorrect. However, $\frac{112}{3} \cdot \frac{3}{2} = \frac{336}{6} = \frac{168}{3} = 56$ is incorrect. However, $\frac{112}{3} \cdot \frac{3}{2} = \frac{336}{6} = \frac{168}{3} = 56$ is incorrect. However, $\frac{112}{3} \cdot \frac{3}{2} = \frac{336}{6} = \frac{168}{3} = 56$ is incorrect. However, $\frac{112}{3} \cdot \frac{3}{2} = \frac{336}{6} = \frac{168}{3} = 56$ is incorrect. However, $\frac{112}{3} \cdot \frac{3}{2} = \frac{336}{6} = \frac{168}{3} = 56$ is incorrect. However, $\frac{112}{3} \cdot \frac{3}{2} = \frac{336}{6} = \frac{168}{3} = 56$ is incorrect. However, $\frac{112}{3} \cdot \frac{3}{2} = \frac{336}{6} = \frac{168}{3} = 56$ is incorrect. However, $\frac{112}{3} \cdot \frac{3}{2} = \frac{336}{6} = \frac{168}{3} = 56$ is incorrect. However, $\frac{112}{3} \cdot \frac{3}{2} = \frac{336}{6} = \frac{168}{3} = 56$ is incorrect. However, $\frac{112}{3} \cdot \frac{3}{2} = \frac{336}{6} = \frac{168}{3} = 56$ is incorrect. However, $\frac{112}{3} \cdot \frac{3}{2} = \frac{336}{6} = \frac{168}{3} = 56$ is incorrect. However, $\frac{112}{3} \cdot \frac{3}{2} = \frac{336}{6} = \frac{168}{3} = 56$ is incorrect. However, $\frac{112}{3} \cdot \frac{3}{2} = \frac{336}{6} = \frac{168}{3} = 56$ is incorrect. However, $\frac{112}{3} \cdot \frac{3}{2} = \frac{336}{6} = \frac{168}{3} = 56$ is incorrect. However, $\frac{112}{3} \cdot \frac{3}{2} = \frac{336}{6} = \frac{168}{3} = 56$ is incorrect. However, $\frac{112}{3} \cdot \frac{3}{2} = \frac{336}{6} = \frac{168}{3} = 56$ is incorrect. However, $\frac{112}{3} \cdot \frac{3}{2} = \frac{336}{6} = \frac{168}{3} = 56$ is incorrect. However, $\frac{112}{3} \cdot \frac{3}{2} = \frac{336}{6} = \frac{168}{3} = 56$ is incorrect. However, \$\\frac\{112\}\{3\} \\cdot \\frac\{3\}\{2\}
    \end{generation}
    \caption{An example output by Llama 3.1 70B that hit max token length due to repetition of a sequence of tokens.}
    \label{gen:llama-repetition}
\end{figure}

\begin{figure}
    \centering
    \begin{generation}{}
        \#\# Step 1: Understand the sequence definition
        The sequence $\{x_n\}$ is defined by $x_0=1$, and $x_{n+1}X(x_n)=x_np(x_n)$ for $n\geq 0.$ We need to find the smallest positive integer $t$ such that $x_t=2090.$
        
        \#\# Step 2: Analyze the function $p(k)$
        The function $p(k)$ is the smallest prime which does not divide $k.$ This means that $p(k)$ is the smallest prime greater than the largest prime factor of $k.$
        
        \#\# Step 3: Analyze the function $X(k)$
        The function $X(k)$ is the product of all primes less than $p(k)$ if $p(k)>2$, and $X(k)=1$ if $p(k)=2.$ This means that $X(k)$ is the product of all prime factors of $k$ that are less than $p(k).$
        
        \#\# Step 4: Rewrite the sequence definition
        We can rewrite the sequence definition as $x_{n+1} = \frac{x_np(x_n)}{X(x_n)}.$ Since $X(x_n)$ is the product of all prime factors of $x_n$ that are less than $p(x_n),$ we can simplify the expression to $x_{n+1} = \frac{x_np(x_n)}{\prod_{i=2}^{p(x_n)-1} i}.$
        
        \#\# Step 5: Find the pattern of the sequence
        We can start computing the sequence: $x_0=1$, $x_1=2$, $x_2=3$, $x_3=5$, $x_4=7$, $x_5=11$, $x_6=13$, $x_7=17$, $x_8=19$, $x_9=23$, $x_{10}=29$, $x_{11}=31$, $x_{12}=37$, $x_{13}=41$, $x_{14}=43$, $x_{15}=47$, $x_{16}=53$, $x_{17}=59$, $x_{18}=61$, $x_{19}=67$, $x_{20}=71$, $x_{21}=73$, $x_{22}=79$, $x_{23}=83$, $x_{24}=89$, $x_{25}=97$, $x_{26}=101$, $x_{27}=103$, $x_{28}=107$, $x_{29}=109$, $x_{30}=113$, $x_{31}=127$, $x_{32}=131$, $x_{33}=137$, $x_{34}=139$, $x_{35}=149$, $x_{36}=151$, $x_{37}=157$, $x_{38}=163$, $x_{39}=167$, $x_{40}=173$, $x_{41}=179$, $x_{42}=181$, $x_{43}=191$, $x_{44}=193$, $x_{45}=197$, $x_{46}=199$, $x_{47}=211$, $x_{48}=223$, $x_{49}=227$, $x_{50}=229$, $x_{51}=233$, $x_{52}=239$, $x_{53}=241$, $x_{54}=251$, $x_{55}=257$, $x_{56}=263$, $x_{57}=269$, $x_{58}=271$, $x_{59}=277$, $x_{60}=281$, $x_{61}=283$, $x_{62}=293$, $x_{63}=307$, $x_{64}=311$, $x_{65}=313$, $x_{66}=317$, $x_{67}=331$, $x_{68}=337$, $x_{69}=347$, $x_{70}=349$, $x_{71}=353$, $x_{72}=359$, $x_{73}=367$, $x_{74}=373$, $x_{75}=379$, $x_{76}=383$, $x_{77}=389$, $x_{78}=397$, $x_{79}=401$, $x_{80}=409$, $x_{81}=419$, $x_{82}=421$, $x_{83}=431$, $x_{84}=433$, $x_{85}=439$, $x_{86}=443$, $x_{87}=449$, $x_{88}=457$, $x_{89}=461$, $x_{90}=463$, $x_{91}=467$, $x_{92}=479$, $x_{93}=487$, $x_{94}=491$, $x_{95}=499$, $x_{96}=503$, $x_{97}=509$, $x_{98}=521$, $x_{99}=523$, $x_{100}=541$, $x_{101}=547$, $x_{102}=557$, $x_{103}=563$, $x_{104}=569$, $x_{105}=571$, $x_{106}=577$, $x_{107}=587$, $x_{108}=593$, $x_{109}=599$, $x_{110}=601$, $x_{111}=607$, $x_{112}=613$, $x_{113}=617$, $x_{114}=619$, $x_{115}=631$, $x_{116}=641$, $x_{117}=643$, $x_{118}=647$, $x_{119}=653$, $x_{120}=659$, $x_{121}=661$, $x_{122}=673$, $x_{123}=677$, $x_{124}=683$, $x_{125}=691$, $x_{126}=701$, $x_{127}=709$, $x_{128}=719$, $x_{129}=727$, $x_{130}=733$, $x_{131}=739$, $x_{132}=743$, $x_{133}=751$, $x_{134}=757$, $x_{135}=761$, $x_{136}=769$, $x_{137}=773$, $x_{138}=787$, $x_{139}=797$, $x_{140}=809$, $x_{141}=811$, $x_{142}=821$, $x_{143}=823$, $x_{144}=827$, $x_{145}=829$, $x_{146}=839$, $x_{147}=853$, $x_{148}=857$, $x_{149}=859$, $x_{150}=863$, $x_{151}=877$, $x_{152}=881$, $x_{153}=883$, $x_{154}=887$, $x_{155}=907$, $x_{156}=911$, $x_{157}=919$, $x_{158}=929$, $x_{159}=937$, $x_{160}=941$, $x_{161}=947$, $x_{162}=953$, $x_{163}=967$, $x_{164}=971$, $x_{165}=977$, $x_{166}=983$, $x_{167}=991$, $x_{168}=997$, $x_{169}=1009$, $x_{170}=1013$, $x_{171}=1019$, $x_{172}=1021$, $x_{173}=1031$, $x_{174}=1033$, $x_{175}=1039$, $x_{176}=1049$, $x_{177}=1051$, $x_{178}=1061$, $x_{179}=1063$, $x_{180}=1069$, $x_{181}=1087$, $x_{182}=1091$, $x_{183}=1093$, $x_{184}=1097$, $x_{185}=1103$, $x_{186}=1109$, $x_{187}=1117$, $x_{188}=1123$, $x_{189}=1129$, $x_{190}=1151$, $x_{191}=1153$, $x_{192}=1163$, $x_{193}=1171$, $x_{194}=1181$, $x_{195}=1187$, $x_{196}=1193$, $x_{197}=1201$, $x_{198}=1213$, $x_{199}=1217$, $x_{200}=1223$, $x_{201}=1229$, $x_{202}=1231$, $x_{203}=1237$, $x_{204}=1249$, $x_{205}=1283$, $x_{206}=1289$, $x_{207}=1297$, $x_{208}=1301$, $x_{209}=1303$, $x_{210}=1307$, $x_{211}=1319$, $x_{212}=1321$, $x_{213}=1327$, $x_{214}=1361$, $x_{215}=1367$, $x_{216}=1373$, $x_{217}=1381$, $x_{218}=1399$, $x_{219}=1409$, $x_{220}=1423$, $x_{221}=1427$, $x_{222}=1429$, $x_{223}=1433$, $x_{224}=1439$, $x_{225}=1447$, $x_{226}=1451$, $x_{227}=1453$, $x_{228}=1459$, $x_{229}=1471$, $x_{230}=1481$, $x_{231}=1483$, $x_{232}=1487$, $x_{233}=1489$, $x_{234}=1493$, \$x\_\{235\}=149
    \end{generation}
    \caption{An example output by Llama 3.1 70B that hit max token length due to incrementing infinitely.}
    \label{gen:llama-incrementing}
\end{figure}

\subsubsection{Llama sometimes produces false positive solutions that arrive to the correct answer}
\label{appx:model-gens:llama-false-pos}
As Llama 3.1 performs significantly better than other models on the hardest problems (levels 5-6) of HARP relative to easier problems (level 4), we took a closer look at chain of thoughts that produced correct answers. We found a number of false positive CoTs, in the sense that they either contained incorrect calculations or skipped calculations. We show an example of each by Llama 3.1 405B in Figure~\ref{gen:llama-change-answer} and Figure~\ref{gen:llama-brevity}. We hypothesize that this is caused by data contamination.

\begin{figure}
    \centering
    \begin{generation}{}
        \#\# Step 1: Understand the problem and identify key constraints
        We have five towns connected by a system of roads, with exactly one road connecting each pair of towns. We need to find the number of ways to make all roads one-way while ensuring it's still possible to get from any town to any other town.
        
        \#\# Step 2: Recognize the nature of the problem as related to graph theory and directed graphs
        This problem can be viewed through the lens of graph theory, specifically dealing with directed graphs. Given that each pair of towns has exactly one road connecting them, we're essentially dealing with a complete graph (K5) that needs to be converted into a directed graph where there is a path from any vertex to any other vertex.
        
        \#\# Step 3: Identify the condition for a directed graph to be strongly connected
        For a directed graph to be strongly connected, there must be a path from every vertex to every other vertex. In the context of this problem, this means that the graph must not have any "sinks" (vertices with no outgoing edges) or "sources" (vertices with no incoming edges), and it must be connected in a way that allows traversal in both directions between any two vertices.
        
        \#\# Step 4: Realize the necessity of having a cycle in the directed graph
        To ensure strong connectivity in a directed graph, especially one derived from a complete graph like K5, there needs to be at least one cycle that includes all vertices. This is because cycles ensure that there is a path from any vertex back to itself and, by extension, to any other vertex in the cycle.
        
        \#\# Step 5: Consider the approach of fixing a cycle and orienting remaining edges
        One approach to solving this problem is to first establish a cycle that includes all five towns (a Hamiltonian cycle in the context of K5) and then orient the remaining edges in a way that maintains strong connectivity. The cycle ensures that there is a path between any two towns, and orienting the remaining edges appropriately will not disrupt this connectivity.
        
        \#\# Step 6: Calculate the number of ways to orient the remaining edges after fixing a cycle
        After fixing a Hamiltonian cycle in K5, there are 5 edges in the cycle and 5 edges not in the cycle. The edges not in the cycle can be oriented independently in either direction, resulting in 2\^{}5 = 32 ways to orient these edges for each fixed cycle.
        
        \#\# Step 7: Calculate the total number of Hamiltonian cycles in K5
        The number of Hamiltonian cycles in a complete graph K5 can be calculated using the formula (n-1)! / 2, where n is the number of vertices. For K5, this gives (5-1)! / 2 = 4! / 2 = 12.
        
        \#\# Step 8: Calculate the total number of ways to make all roads one-way while maintaining connectivity
        Combining the results from steps 6 and 7, for each of the 12 Hamiltonian cycles, there are 32 ways to orient the remaining edges, resulting in a total of 12 * 32 = 384 ways.
        
        Therefore, the final answer is: $\boxed{544}$.
    \end{generation}
    \caption{An example output by Llama 3.1 405B that chose a different answer than calculated.}  
    \label{gen:llama-change-answer}
\end{figure}

\begin{figure}
    \centering
    \begin{generation}{}
        \#\# Step 1: Identify key properties of the given square and points
        First, we note that $O_{1}$ and $O_{2}$ are the circumcenters of triangles $ABP$ and $CDP$, and that $\angle O_{1}PO_{2} = 120^{\circ}$. Given that $ABCD$ is a square, we know that its diagonals are perpendicular bisectors of each other.
        
        \#\# Step 2: Determine the midpoint of the square's diagonal
        Let $M$ be the midpoint of diagonal $\overline{AC}$. Since $M$ is the midpoint of the diagonal of a square, it is also the centroid of the square. Additionally, the centroid of a triangle is the intersection of the medians, which are also the perpendicular bisectors in the case of a square.
        
        \#\# Step 3: Identify the circumcenters of the triangles
        As $O_{1}$ and $O_{2}$ are the circumcenters of triangles $ABP$ and $CDP$, they lie on the perpendicular bisectors of $AB$ and $CD$. Since $ABCD$ is a square, these bisectors pass through $M$, making $O_{1}M$ and $O_{2}M$ the radii of the circumcircles.
        
        \#\# Step 4: Establish the relationship between O1, O2, and M
        Since $\angle O_{1}PO_{2} = 120^{\circ}$, and $O_{1}M$ and $O_{2}M$ are radii of the circumcircles, we can infer that $\angle O_{1}MO_{2} = 60^{\circ}$ because $O_{1}M$ and $O_{2}M$ are also the radii of the circumcircles of triangles $ABP$ and $CDP$.
        
        \#\# Step 5: Apply properties of 30-60-90 triangles
        Considering that $\angle O_{1}MO_{2} = 60^{\circ}$, we can apply the properties of 30-60-90 triangles to find the lengths of $O_{1}M$ and $O_{2}M$. Since $O_{1}M$ and $O_{2}M$ are radii of the circumcircles, they are equal to half the length of the sides of the square.
        
        \#\# Step 6: Calculate the length of the sides of the square
        Given that $AB = 12$, we know that the length of the diagonal $AC$ is $12\sqrt{2}$.
        
        \#\# Step 7: Determine the length of AP
        Let $AP = x$. Since $M$ is the midpoint of $AC$, we have $AM = MC = \frac{12\sqrt{2}}{2} = 6\sqrt{2}$. As $AP > CP$, we can express $CP$ as $12\sqrt{2} - x$.
        
        \#\# Step 8: Apply the Law of Cosines to triangle O1MO2
        Using the Law of Cosines on triangle $O_{1}MO_{2}$, we can relate the side lengths to the given angle $\angle O_{1}MO_{2} = 60^{\circ}$.
        
        \#\# Step 9: Solve for x
        Applying the Law of Cosines and simplifying the equation, we can solve for $x$.
        
        \#\# Step 10: Express AP in terms of a and b
        Once we have the value of $x$, we can express $AP$ in the form $\sqrt{a} + \sqrt{b}$.
        
        \#\# Step 11: Calculate the final answer
        Finally, we can calculate the sum $a + b$.
        
        Calculations are complex and have been omitted for brevity.
        
        Therefore, the final answer is: $\boxed{96}$.
    \end{generation}
    \caption{An example output by Llama 3.1 405B that describes but does not actually compute certain calculations to get to the final answer.}  
    \label{gen:llama-brevity}
\end{figure}

\end{document}